\documentclass[lettersize,journal]{IEEEtran}
\usepackage{amsmath,amsfonts}
\usepackage{algorithmic}
\usepackage{algorithm}
\usepackage{array}
\usepackage[caption=false,font=normalsize,labelfont=sf,textfont=sf]{subfig}
\usepackage{textcomp}
\usepackage{stfloats}
\usepackage{url}
\usepackage{verbatim}
\usepackage{graphicx}
\usepackage{threeparttable}
\usepackage{cite}
\usepackage{tikz}
\usepackage{pgfplots}
\usepackage[font=small,labelfont=bf,justification=centering]{caption}
\usetikzlibrary{shapes.geometric, arrows, positioning, arrows.meta, calc, patterns, decorations.pathreplacing, backgrounds, fit, matrix, shapes.symbols, shapes.callouts, shapes.multipart, shapes.gates.logic.US, shapes.gates.logic.IEC, er, automata, topaths, trees, petri, mindmap, calendar, external, fadings, through, intersections, pgfplots.groupplots, pgfplots.fillbetween, pgfplots.polar, pgfplots.smithchart, pgfplots.statistics, pgfplots.colorbrewer, pgfplots.patchplots, pgfplots.units, pgfplots.dateplot, pgfplots.polar, pgfplots.smithchart, pgfplots.statistics, pgfplots.colorbrewer, pgfplots.patchplots, pgfplots.units, pgfplots.dateplot}
\usepackage{subcaption}

\pgfplotsset{compat=1.18}
\begin{document}

\title{Image Segmentation with Large Language Models: A Survey with Perspectives for Intelligent Transportation Systems}

\author{%
  Sanjeda Akter\IEEEauthorrefmark{1},
  Ibne Farabi Shihab\IEEEauthorrefmark{1},
  Anuj Sharma\IEEEauthorrefmark{2}%
  \thanks{\IEEEauthorrefmark{1}Department of Computer Science, Iowa State University, Ames, IA, USA
          (e-mail: sanjeda@iastate.edu; ishihab@iastate.edu).}%
  \thanks{\IEEEauthorrefmark{2}Department of Civil, Construction and Environmental Engineering,
          Iowa State University, Ames, IA, USA.}%
  \thanks{Sanjeda Akter and Ibne Farabi Shihab contributed equally to this work.}%
}

\markboth{IEEE Transactions on Intelligent Transportation Systems,~Vol.~XX, No.~X, Month~Year}%
{Author \MakeLowercase{\textit{et al.}}: Image Segmentation with Large Language Models}

\IEEEpubid{0000--0000/00\$00.00~\copyright~2025 IEEE}
\IEEEpubidadjcol

\maketitle

\begin{abstract}
The integration of Large Language Models (LLMs) with computer vision is profoundly transforming perception tasks like image segmentation. For intelligent transportation systems (ITS), where accurate scene understanding is critical for safety and efficiency, this new paradigm offers unprecedented capabilities. This survey systematically reviews the emerging field of LLM-augmented image segmentation, focusing on its applications, challenges, and future directions within ITS. We provide a taxonomy of current approaches based on their prompting mechanisms and core architectures, and we highlight how these innovations can enhance road scene understanding for autonomous driving, traffic monitoring, and infrastructure maintenance. Finally, we identify key challenges, including real-time performance and safety-critical reliability, and outline a perspective centered on explainable, human-centric AI as a prerequisite for the successful deployment of this technology in next-generation transportation systems.
\end{abstract}


%Follow this paper as a template kind of : https://ieeexplore.ieee.org/stamp/stamp.jsp?tp=&arnumber=10265760

\begin{figure*}[htbp]
    \centering
    \includegraphics[width=1.0\textwidth]{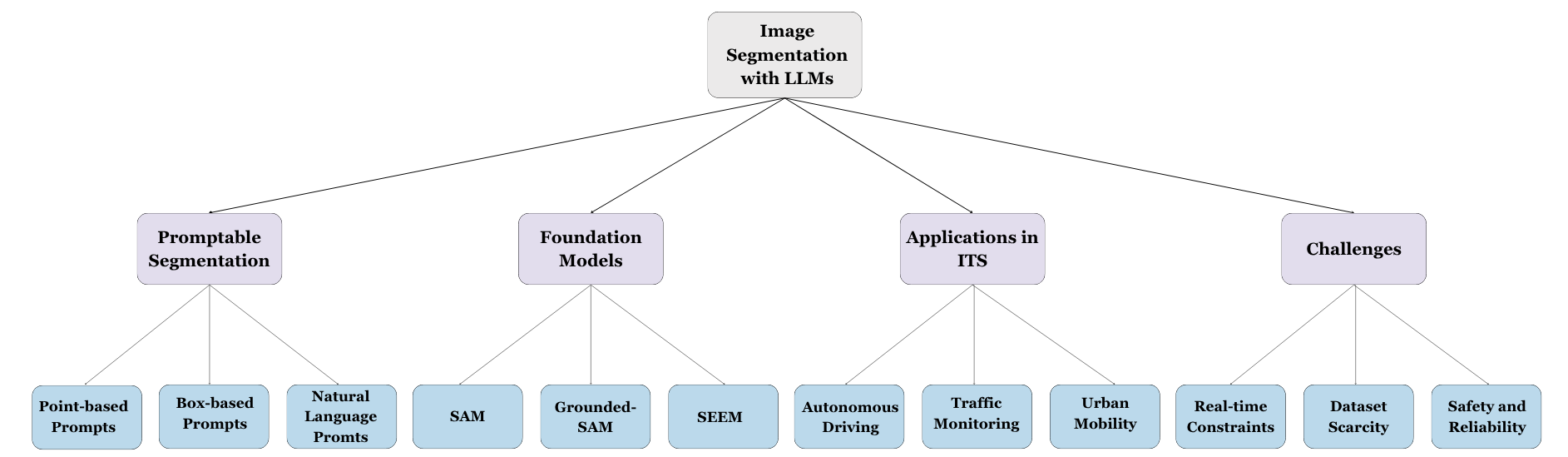}  % Change width and filename
    \caption{Taxonomy of Image Segmentation with Large Language Models for Intelligent Transportation Systems}
    \label{fig:taxonomy}
\end{figure*}

\section{Introduction}
\label{sec: intro}

\IEEEPARstart{I}{mage} segmentation is a foundational component of visual perception in intelligent transportation systems (ITS), enabling autonomous vehicles to interpret complex driving environments with precision and reliability \cite{zhou2024visionlanguagemodelsautonomous}. By delineating road lanes, detecting obstacles, segmenting pedestrians, and recognizing traffic signs, segmentation empowers vehicles to navigate urban and rural settings safely and efficiently \cite{huang2024multimodalsensorfusionauto}. Historically, segmentation tasks have relied on convolutional neural networks (CNNs), such as DeepLab \cite{chen2017deeplab} and Mask R-CNN \cite{he2017mask}, and more recently on vision-specific transformers, such as Swin Transformer \cite{liu2021swin} and Segmenter \cite{strudel2021segmenter}, which have achieved remarkable performance on benchmark datasets like Cityscapes \cite{cordts2016cityscapesdatasetsemanticurban} and BDD100K \cite{yu2020bdd100k}. However, the emergence of Large Language Models (LLMs) has introduced a paradigm shift, leveraging their advanced language understanding and reasoning capabilities to enhance image segmentation through multimodal learning \cite{zhou2024visionlanguagemodelsautonomous, huang2024multimodalsensorfusionauto}. This survey explores the convergence of LLMs and image segmentation, with a particular focus on their transformative applications in ITS.

Image segmentation in ITS encompasses a range of tasks critical to autonomous driving and traffic management. Semantic segmentation assigns class labels to each pixel, enabling the identification of road surfaces, vehicles, and pedestrians \cite{chen2018deeplabv3plus}. Instance segmentation distinguishes individual objects within the same class, such as separating multiple pedestrians in a crowd \cite{he2017mask}. Panoptic segmentation combines both approaches to provide a holistic scene understanding, crucial for complex urban environments \cite{kirillov2019panoptic}. These tasks have traditionally been driven by CNN-based architectures, which excel at capturing spatial hierarchies but often require extensive labeled datasets and struggle with open-vocabulary scenarios \cite{guo2020semantic}. Vision transformers, such as the Vision Transformer (ViT) \cite{dosovitskiy2020vit}, have addressed some limitations by leveraging self-attention mechanisms to model long-range dependencies, improving performance on datasets like nuScenes \cite{caesar2020nuscenes} and Mapillary Vistas \cite{neuhold2017mapillary}. Yet, these models remain constrained by their reliance on predefined class labels and limited adaptability to dynamic or novel scenarios \cite{zhou2024visionlanguagemodelsautonomous}.

The integration of LLMs into image segmentation, often referred to as vision-language segmentation (VLSeg), represents a significant leap forward. LLMs, such as BERT \cite{devlin2018bert}, GPT-3 \cite{brown2020gpt3}, and T5 \cite{raffel2020t5}, are renowned for their ability to understand and generate human-like text, enabling them to process natural language prompts for guiding segmentation tasks \cite{zhou2024visionlanguagemodelsautonomous}. By combining LLMs with vision models like CLIP \cite{radford2021clip} or DINOv2 \cite{oquab2023dinov2}, VLSeg frameworks, such as Grounded-SAM \cite{liu2023groundedsam} and SEEM \cite{zou2023seem}, allow autonomous systems to segment objects based on free-form queries, such as ``highlight the cyclist on the right'' or ``segment the traffic cone near the construction zone'' \cite{zou2023seem, liu2023groundedsam}. This flexibility is particularly valuable in ITS, where vehicles must adapt to diverse and unpredictable environments, including adverse weather, occlusions, or novel obstacles \cite{huang2024multimodalsensorfusionauto, dalcol2024jointperceptionpredictionautonomous}.

In ITS, LLM-augmented VLSeg has transformative potential across several applications. For autonomous driving, it enables real-time scene parsing, dynamic obstacle detection, and predictive segmentation of crash scenarios, as demonstrated by frameworks like DriveLM \cite{sima2024drivelm} and InsightGPT \cite{chen2024insightgpt}. For traffic monitoring, VLSeg supports smarter traffic flow analysis and anomaly detection in surveillance feeds, enhancing city-scale mobility solutions \cite{li2024llm4tr, wang2023traffic}. Datasets like Talk2Car \cite{deruytter2019talk2car} and Road-Seg-VL \cite{road_seg_vl_2024} provide language-guided annotations, facilitating the development of models that align visual perception with human instructions \cite{zhou2024visionlanguagemodelsautonomous}. Moreover, recent advancements in open-vocabulary segmentation, such as CLIPSeg \cite{luddecke2022clipseg} and OpenSeg \cite{ghiasi2022openseg}, enable zero-shot segmentation of unseen objects, addressing the open-world challenges inherent in ITS \cite{ghiasi2022openseg}. A central theme of this survey is the critical trade-off we term the 'cost of generalization': while powerful VLSeg models offer unprecedented flexibility for open-world scenarios, they often lag in performance on specific, safety-critical tasks compared to highly optimized, specialized models. Analyzing this trade-off is essential for their practical deployment in ITS.

Despite these advancements, integrating LLMs into segmentation for ITS faces several challenges. Real-time performance is a critical bottleneck, as large models like SAM \cite{kirillov2023segment} incur high computational costs, necessitating lightweight solutions like MobileSAM \cite{zhang2023mobilesam} or EdgeViT \cite{pan2022edgevitscompetinglightweightcnns}. Reliability in safety-critical scenarios requires robustness against adversarial inputs and adverse conditions, as explored in Multi-Shield \cite{wang2023robust}. Dataset limitations, particularly the scarcity of large-scale multimodal datasets, hinder model training, though automated annotation pipelines like AutoSeg \cite{zhao2024autoseg} offer promising solutions \cite{zhao2024autoseg, huang2024multimodalsensorfusionauto}. This survey aims to provide a comprehensive analysis of these developments, challenges, and future directions, offering insights into how LLM-augmented VLSeg can reshape ITS to be safer, more adaptable, and intelligent \cite{zhou2024visionlanguagemodelsautonomous, huang2024multimodalsensorfusionauto}.

The remainder of this paper is organized as follows: Section 2 reviews LLMs and promptable segmentation techniques, Section 3 discusses their applications in ITS, Section 4 examines relevant datasets and benchmarks, Section 5 addresses key challenges, Section 6 explores future directions, and Section 7 concludes with a synthesis of findings and prospects for LLM-augmented VLSeg in ITS.

\section{Background and Related Works}
\label{sec:background}

\subsection{Image Segmentation Fundamentals}
\label{sub:imgseg}
Image segmentation is a pivotal task in computer vision that involves partitioning an image into distinct segments or regions, each representing meaningful objects or areas. In the context of intelligent transportation systems (ITS), image segmentation underpins the ability of autonomous vehicles to interpret complex driving environments by identifying road lanes, detecting obstacles, recognizing pedestrians, and understanding traffic signs \cite{driving_forward_2024}. By providing a structured representation of visual scenes, segmentation enables critical functionalities such as scene understanding, path planning, collision avoidance, and urban mobility management. The task is broadly categorized into three primary types, each serving distinct purposes in ITS applications \cite{elhassan2024real, kirillov2019panoptic}.

\begin{itemize}
    \item \textbf{Semantic Segmentation}: This approach assigns a class label to each pixel in an image, enabling the identification of categories such as roads, vehicles, pedestrians, and traffic signs. Semantic segmentation is essential in ITS for delineating drivable areas, distinguishing road boundaries, and understanding the overall structure of a driving scene \cite{elhassan2024real}. Models like DeepLab \cite{chen2017deeplab} and its successors, such as DeepLabv3+ \cite{chen2018deeplabv3plus}, leverage atrous convolutions and spatial pyramid pooling to capture multi-scale contextual information, achieving high accuracy on urban scene datasets like Cityscapes \cite{cordts2016cityscapesdatasetsemanticurban}. Semantic segmentation is particularly valuable for tasks like road surface detection and traffic signal recognition, ensuring safe navigation in diverse environments.

    \item \textbf{Instance Segmentation}: Unlike semantic segmentation, instance segmentation differentiates individual objects within the same class, such as separating multiple pedestrians or vehicles in a crowded urban scene. This granularity is crucial for autonomous driving, where precise localization of individual entities is necessary for path planning, obstacle avoidance, and interaction with dynamic objects. Mask R-CNN \cite{he2017mask}, built upon Faster R-CNN \cite{ren2015faster}, is a cornerstone model that combines object detection with pixel-wise segmentation, enabling instance-level precision in ITS applications \cite{vehicle_detection_2023}. Recent advancements, such as Mask2Former \cite{cheng2022mask2former}, further enhance instance segmentation by integrating transformer-based architectures, improving performance on datasets like BDD100K \cite{yu2020bdd100k}.

    \item \textbf{Panoptic Segmentation}: Panoptic segmentation unifies semantic and instance segmentation by assigning both class labels and instance IDs to every pixel, providing a comprehensive scene representation. This holistic approach is particularly valuable in complex ITS scenarios, where autonomous vehicles must navigate urban environments with diverse, interacting elements, such as pedestrians, vehicles, and infrastructure \cite{kirillov2019panoptic}. Models like OneFormer \cite{jain2022oneformer} leverage transformer architectures to achieve state-of-the-art panoptic segmentation, excelling on datasets like nuScenes \cite{caesar2020nuscenes} and Mapillary Vistas \cite{neuhold2017mapillary}. Panoptic segmentation supports advanced scene understanding, enabling vehicles to reason about both static and dynamic elements in real-time \cite{elhassan2024real}.
\end{itemize}

\subsection{Classical Approaches to Image Segmentation}
\label{sub:classical}
The journey of image segmentation in ITS began with classical computer vision techniques, such as thresholding and region-growing methods, but the field was fundamentally transformed by the advent of deep learning and Convolutional Neural Networks (CNNs). These classical approaches serve as the bedrock upon which modern, more sophisticated models are built.

\subsubsection{Historical Evolution of Segmentation Approaches}
\label{subsubsec:historical_evolution}

\begin{figure*}[h]
 \centering   
 \includegraphics[width = \textwidth]{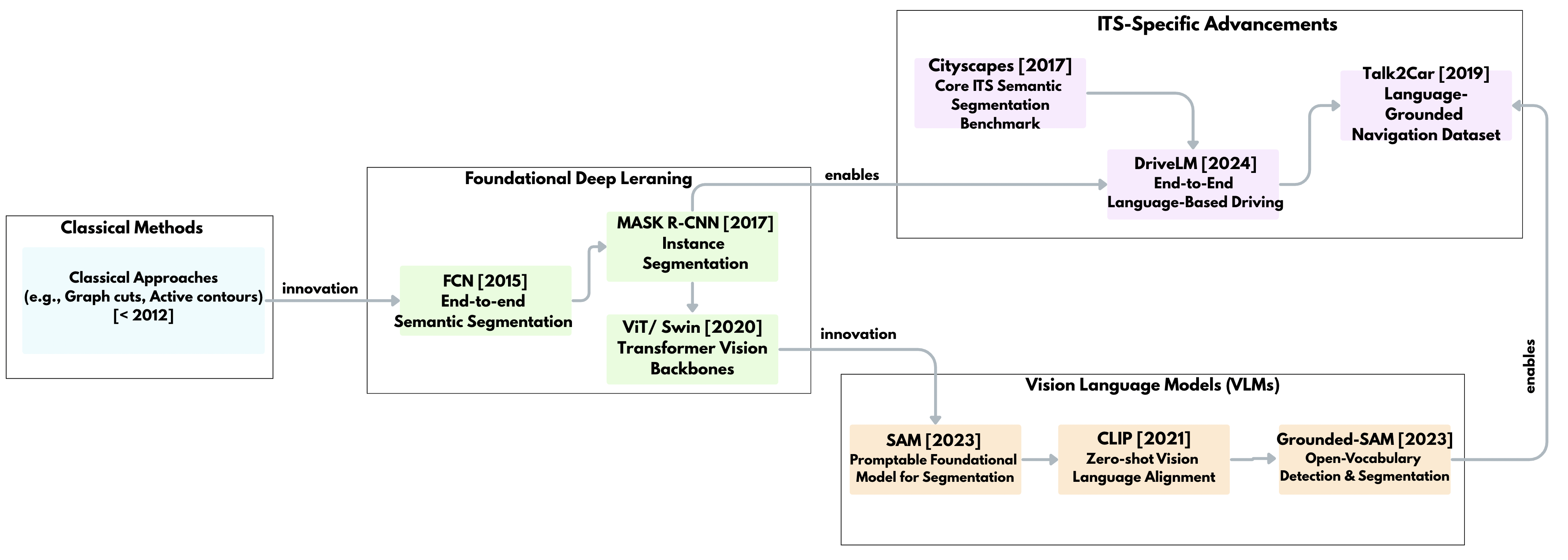}
 \caption{Timeline of key milestones in the evolution of segmentation for ITS, with a color-coded legend at the top indicating categories.}
    \label{fig:its_vlseg_timeline}
\end{figure*}

The evolution of segmentation methods for ITS applications can be traced through several distinct phases, each building upon the innovations of the previous era:

\begin{itemize}
    \item \textbf{Pre-Deep Learning Era (1980s-2012):} Early approaches to segmentation relied on classical computer vision techniques. These included:
    \begin{itemize}
        \item \textbf{Edge-Based Methods:} Algorithms like Canny edge detection identified object boundaries based on intensity gradients.
        \item \textbf{Region-Based Methods:} Techniques such as region growing, watershed algorithms, and mean-shift clustering grouped pixels based on homogeneity criteria.
        \item \textbf{Graph-Based Methods:} Approaches like Normalized Cuts \cite{shi2000normalized} and Graph Cuts formulated segmentation as a graph partitioning problem.
        \item \textbf{Model-Based Methods:} Active contours (snakes) and level sets evolved contours to fit object boundaries.
    \end{itemize}
    
    These classical methods were computationally efficient but struggled with the semantic understanding required for complex driving scenes. They typically relied on low-level features (color, texture, edges) and required careful parameter tuning for each specific scenario, making them brittle in the face of real-world variability.
    
    \item \textbf{Early Deep Learning Era (2012-2015):} The breakthrough came with the application of CNNs to segmentation tasks:
    \begin{itemize}
        \item \textbf{Patch Classification:} Early approaches treated segmentation as a per-pixel classification problem, where a CNN classified the central pixel of each image patch. While effective, this was computationally inefficient.
        \item \textbf{Fully Convolutional Networks (FCN) \cite{long2015fcn}:} The seminal work by Long et al. in 2015 replaced fully connected layers with convolutional layers, enabling end-to-end training for dense prediction. FCNs could process arbitrary-sized inputs and produce correspondingly-sized outputs, dramatically improving efficiency.
        \item \textbf{Early Encoder-Decoder Architectures:} Models like SegNet \cite{badrinarayanan2017segnet} and U-Net \cite{ronneberger2015unet} introduced the encoder-decoder paradigm, where an encoder network downsamples the input to capture context, and a decoder upsamples to recover spatial details.
    \end{itemize}
    
    \item \textbf{Refinement Era (2016-2019):} This period saw significant architectural innovations to address the limitations of early deep learning approaches:
    \begin{itemize}
        \item \textbf{Multi-Scale Processing:} DeepLab \cite{chen2017deeplab} introduced atrous (dilated) convolutions and Atrous Spatial Pyramid Pooling (ASPP) to capture multi-scale context without increasing computational cost.
        \item \textbf{Attention Mechanisms:} Models began incorporating spatial and channel attention to focus on the most informative regions and features.
        \item \textbf{Instance-Level Reasoning:} Mask R-CNN \cite{he2017mask} extended Faster R-CNN \cite{ren2015faster} by adding a branch for predicting segmentation masks, enabling instance segmentation.
        \item \textbf{Panoptic Segmentation:} Kirillov et al. \cite{kirillov2019panoptic} introduced the concept of panoptic segmentation, unifying semantic and instance segmentation into a single task.
    \end{itemize}
    
    \item \textbf{Transformer Era (2020-2021):} The introduction of transformers to computer vision marked another paradigm shift:
    \begin{itemize}
        \item \textbf{Vision Transformer (ViT) \cite{dosovitskiy2020vit}:} By treating an image as a sequence of patches, ViT applied self-attention mechanisms to model global relationships, though it was initially designed for image classification rather than segmentation.
        \item \textbf{Hierarchical Transformers:} Models like Swin Transformer \cite{liu2021swin} addressed the limitations of ViT for dense prediction tasks by introducing a hierarchical structure with local attention windows.
        \item \textbf{Transformer-Based Segmentation:} Architectures like Segmenter \cite{strudel2021segmenter}, SegFormer \cite{xie2021segformer}, and Mask2Former \cite{cheng2022mask2former} adapted transformers specifically for segmentation tasks, achieving state-of-the-art performance.
    \end{itemize}
    
    \item \textbf{Early Vision-Language Models (2019-2021):} Before the current era of LLM-augmented segmentation, several approaches attempted to bridge vision and language for segmentation:
    \begin{itemize}
        \item \textbf{Referring Expression Segmentation:} Models like CMSA \cite{ye2019cross} and BRINet \cite{hu2020bi} focused on segmenting objects referred to by natural language expressions, typically using LSTMs or GRUs to encode text.
        \item \textbf{Visual Grounding:} Approaches like PhraseCut \cite{wu2020phrasecut} and RefVOS \cite{bellver2020refvos} aimed to localize and segment objects based on natural language descriptions.
        \item \textbf{Vision-Language Pre-training:} Models like ViLBERT \cite{lu2019vilbert} and LXMERT \cite{tan2019lxmert} performed joint pre-training of vision and language representations, though not specifically for segmentation.
    \end{itemize}
    These early approaches laid important groundwork but were limited by their reliance on relatively simple language models and task-specific training.
\end{itemize}

The first major breakthrough in deep learning-based segmentation was the Fully Convolutional Network (FCN) \cite{long2015fcn}, which replaced the dense, fully-connected layers of classification networks (like AlexNet \cite{krizhevsky2012alexnet}) with convolutional layers, enabling end-to-end training for pixel-wise prediction. For ITS, this meant that models could now learn to segment entire driving scenes at once. Following this, architectures like SegNet \cite{badrinarayanan2017segnet} and U-Net \cite{ronneberger2015unet} introduced the powerful encoder-decoder paradigm. The encoder, typically a pre-trained classification network (e.g., VGG \cite{simonyan2014vgg}), progressively downsamples the input to capture semantic context, while the decoder upsamples these features to reconstruct a full-resolution segmentation map. U-Net's key innovation was the use of ``skip connections," which pass fine-grained feature details from the encoder directly to the decoder, proving crucial for accurately localizing small objects like traffic signs and pedestrians in ITS scenes. To handle the vast variation in object scales in driving environments (e.g., distant cars vs. nearby trucks), the DeepLab family of models \cite{chen2017deeplab, chen2018deeplabv3plus} introduced atrous (or dilated) convolutions and Atrous Spatial Pyramid Pooling (ASPP), allowing the network to probe features at multiple resolutions without increasing computational cost. These CNN-based models became the dominant approach and achieved impressive results on benchmarks like Cityscapes \cite{cordts2016cityscapesdatasetsemanticurban}.

The next paradigm shift was the introduction of the Vision Transformer (ViT) \cite{dosovitskiy2020vit}. Inspired by the success of transformers in natural language processing \cite{vaswani2023attentionneed}, ViTs treat an image as a sequence of patches and use self-attention mechanisms to model global relationships between them. This was a departure from the localized receptive fields of CNNs. For complex urban environments in ITS, self-attention provided a mechanism to model long-range dependencies, for example, understanding the relationship between a traffic light on one side of the road and the lane markings on the other. The Swin Transformer \cite{liu2021swin} made ViTs more efficient and effective for dense vision tasks by introducing a hierarchical structure and computing self-attention within shifted windows. Models like Segmenter \cite{strudel2021segmenter} and SegFormer \cite{xie2021segformer} further adapted the transformer architecture specifically for semantic segmentation, demonstrating state-of-the-art performance. These powerful CNN and transformer backbones remain integral components of the more advanced LLM-augmented systems discussed in this survey.

\subsubsection{The CLIP Revolution and Its Impact on Segmentation}
\label{subsubsec:clip_revolution}

The introduction of Contrastive Language-Image Pre-training (CLIP) \cite{radford2021clip} by OpenAI in 2021 marked a pivotal moment in the evolution of vision-language models. CLIP's key innovation was its training methodology: rather than training on a specific task with labeled data, it was trained on 400 million image-text pairs collected from the internet, learning to align images and their textual descriptions in a shared embedding space. This approach enabled zero-shot transfer to a wide range of vision tasks without task-specific fine-tuning.

The impact of CLIP on segmentation was profound and multi-faceted:

\begin{itemize}
    \item \textbf{Open-Vocabulary Capability:} Prior to CLIP, segmentation models were limited to a closed set of predefined classes. CLIP enabled models to segment arbitrary objects described in natural language, dramatically expanding the range of objects that could be identified in driving scenes.
    
    \item \textbf{Bridging Vision and Language:} CLIP provided a natural bridge between visual and linguistic understanding, enabling more intuitive interfaces for segmentation systems. Instead of being constrained to a fixed ontology, users could now query the system using natural language.
    
    \item \textbf{Foundation for VLSeg:} CLIP's architecture and pre-trained weights became the foundation for numerous VLSeg models. CLIPSeg \cite{luddecke2022clipseg}, one of the earliest CLIP-based segmentation models, demonstrated that CLIP's embeddings could be effectively adapted for dense prediction tasks with minimal additional training.
    
    \item \textbf{Compositional Understanding:} CLIP's exposure to diverse image-text pairs enabled a degree of compositional understanding, allowing models to reason about object attributes (color, size, position) and relationships—a critical capability for complex driving scenes.
\end{itemize}

The progression from CLIP to modern VLSeg models followed several key developments:

\begin{enumerate}
    \item \textbf{Direct Adaptation:} Early approaches like CLIPSeg \cite{luddecke2022clipseg} directly adapted CLIP's embeddings for segmentation by adding a lightweight decoder.
    
    \item \textbf{Hybrid Approaches:} Models like OpenSeg \cite{ghiasi2022openseg} combined CLIP's open-vocabulary capabilities with traditional segmentation architectures, using CLIP to generate class embeddings that were then matched with pixel-level features.
    
    \item \textbf{Foundation Models:} The Segment Anything Model (SAM) \cite{kirillov2023segment} introduced a new paradigm of promptable segmentation, trained on a massive dataset of 11 million images and 1.1 billion masks. While SAM itself used geometric prompts rather than language, it laid the groundwork for language-guided segmentation.
    
    \item \textbf{Integrated Approaches:} Models like Grounded-SAM \cite{liu2023groundedsam} and SEEM \cite{zou2023seem} integrated CLIP-based language understanding with SAM's powerful segmentation capabilities, creating systems that could segment objects based on complex natural language descriptions.
\end{enumerate}

This historical progression reveals how the field has evolved from simple, rule-based approaches to sophisticated, language-guided systems capable of understanding and segmenting complex driving scenes based on natural language instructions. The integration of LLMs represents the latest chapter in this evolution, further enhancing the semantic understanding and reasoning capabilities of segmentation models.

Despite their success, these classical approaches face significant limitations in ITS. They typically rely on a closed set of predefined class labels, making them unable to adapt to open-vocabulary scenarios where novel objects, such as temporary road barriers or debris, must be segmented \cite{guo2020semantic}. Moreover, creating the large-scale, pixel-perfect labeled datasets required for training is incredibly costly and time-consuming for diverse ITS scenarios \cite{semantic_segmentation_datasets_2023}. Real-time performance is another persistent challenge, as many of these models are computationally intensive \cite{elhassan2024real}. The integration of Large Language Models (LLMs), as explored in subsequent sections, directly addresses these limitations by enabling context-aware, instruction-guided, and open-vocabulary segmentation, significantly enhancing the adaptability and intelligence of ITS applications \cite{yu2024llmseg, cui2023surveymultimodallargelanguage}.

\subsection{Vision-Language Segmentation}
\label{sub:vls}
Vision-language segmentation (VLSeg) enables the delineation of specific objects or regions within an image based on free-form natural language prompts, offering a significant advancement over traditional segmentation tasks that rely on predefined class labels. In ITS, VLSeg allows autonomous vehicles to respond to dynamic queries, such as ``segment the broken stop sign'' or ``highlight the bus stop next to the corner,'' enhancing adaptability in unpredictable driving scenarios \cite{zou2023seem, yu2024llmseg}. This modality fusion, combining visual and linguistic information, has gained traction with the development of large-scale vision-language models (VLMs) and segmentation foundation models, providing semantic richness, context-awareness, and flexibility critical for ITS applications \cite{radford2021clip, li2023foundation}.

The VLSeg pipeline typically comprises two core modules: (1) a visual encoder that extracts dense spatial features from input images, often using transformer-based backbones like ViTs \cite{dosovitskiy2020vit} or Swin Transformers \cite{liu2021swin}, and (2) a language encoder that translates natural language prompts into feature embeddings, commonly leveraging LLMs such as CLIP's text encoder \cite{radford2021clip} or BERT \cite{devlin2018bert}. Multi-modal fusion modules, typically transformer-based, align visual and linguistic features to predict fine-grained segmentation masks \cite{liu2023groundedsam}. Recent architectures, as illustrated in modern VLSeg frameworks, employ task-specific heads to generate binary or multi-class masks aligned with the provided language description, supporting diverse prompting strategies like text, points, or bounding boxes \cite{zou2023seem, chen2024vlsegarch}.

Key advancements in VLSeg have been driven by foundation models like the Segment Anything Model (SAM) \cite{kirillov2023segment}, which generates high-quality segmentation masks from minimal prompts, and its language-augmented variants, such as Grounded-SAM \cite{liu2023groundedsam} and SEEM (Segment Everything Everywhere Model) \cite{zou2023seem}. Grounded-SAM integrates SAM with grounding techniques to support open-vocabulary queries, enabling zero-shot segmentation of ITS-specific objects like ``traffic cones near the intersection'' \cite{liu2023groundedsam}. SEEM's versatile prompting capabilities make it suitable for dynamic driving scenarios, allowing segmentation based on complex instructions \cite{zou2023seem}. Additionally, video-based VLSeg models like XMem \cite{cheng2022xmemlongtermvideoobject} enable temporal consistency in object tracking, critical for monitoring moving objects like vehicles or cyclists in ITS \cite{cheng2022xmemlongtermvideoobject}. Models like CLIPSeg \cite{luddecke2022clipseg} and OpenSeg \cite{ghiasi2022openseg} further enhance open-vocabulary segmentation, addressing long-tail classes such as rare traffic signs or unexpected obstacles \cite{ghiasi2022openseg, zhang2024openvls}.

Despite these advancements, deploying VLSeg in safety-critical ITS applications presents challenges, including latency constraints for real-time processing, robustness against occlusions, and generalization to rare or unseen objects \cite{yu2024llmseg}. Recent work on lightweight models like MobileSAM \cite{zhang2023mobilesam} and EdgeViT \cite{pan2022edgevitscompetinglightweightcnns} aims to address latency issues, while frameworks like DriveLM \cite{sima2024drivelm} enhance contextual reasoning for complex driving scenarios \cite{sima2024drivelm, zhang2024vlmad}. These developments underscore the potential of VLSeg to transform ITS by enabling intelligent, adaptive, and context-aware scene understanding.

\subsection{Related Work and Contributions}

\begin{figure*}[h]
    \centering
    
    \label{fig:hierarchical_categorization}
    \begin{tikzpicture}[
        node distance=0.7cm and 1.5cm,
        level/.style={
            rectangle, 
            draw, 
            rounded corners,
            fill=blue!10, 
            text width=5cm, 
            align=center, 
            minimum height=1.2cm,
            font=\bfseries
        },
        item/.style={
            rectangle,
            draw,
            fill=gray!10,
            text width=5cm,
            align=center,
            minimum height=1.2cm
        },
        arrow/.style={-stealth, thick, draw=black!80, rounded corners}
    ]
    % Define Level Titles
    \node (low) [level] {Low-Level Research \\ \small (Core Components \& Concepts)};
    \node (mid) [level, right=of low] {Mid-Level Research \\ \small (Task-Oriented VLSeg Models)};
    \node (high) [level, right=of mid] {High-Level Research \\ \small (Integrated Systems \& Applications)};
    
    % Items under Low-Level
    \node (low1) [item, below=of low] {Foundational Algorithms \\ \small (e.g., FCN, U-Net, DeepLab)};
    \node (low2) [item, below=of low1] {Vision Backbones \\ \small (e.g., ViT, Swin Transformer)};
    \node (low3) [item, below=of low2] {V-L Pre-training \\ \small (e.g., CLIP, DINOv2)};

    % Items under Mid-Level
    \node (mid1) [item, below=of mid] {Zero-Shot \& Open-Vocabulary \\ \small (e.g., CLIPSeg, OpenSeg)};
    \node (mid2) [item, below=of mid1] {Promptable Foundation Models \\ \small (e.g., SAM)};
    \node (mid3) [item, below=of mid2] {Hybrid \& Interactive Models \\ \small (e.g., Grounded-SAM, SEEM)};

    % Items under High-Level
    \node (high1) [item, below=of high] {End-to-End Driving \\ \small (e.g., DriveLM, LMDrive)};
    \node (high2) [item, below=of high1] {Traffic Management \\ \small (e.g., TrafficGPT, LLM4Tr)};
    \node (high3) [item, below=of high2] {Human-AI Collaboration \\ \small (e.g., Talk2BEV, Voyager)};

    % Draw arrows between levels
    \draw[arrow] (low.east) -- (mid.west);
    \draw[arrow] (mid.east) -- (high.west);
    
    % Draw arrows to show dependencies
    \coordinate[below = 0.5cm of low.south] (low_out);
    \coordinate[below = 0.5cm of mid.south] (mid_out);
    
    \draw[arrow, gray!60, dashed] (low_out) -| ++(0, -3.5cm) -| (mid1);
    \draw[arrow, gray!60, dashed] (mid_out) -| ++(0, -3.5cm) -| (high1);

    \end{tikzpicture}
    \caption{A hierarchical categorization of the reviewed literature. Research flows from low-level foundational concepts to mid-level task-specific models, and finally to high-level integrated systems for ITS applications.}
\end{figure*}

While this survey provides a broad overview of VLSeg for ITS, it builds upon several recent, more focused surveys. Zhou et al. \cite{zhou2024visionlanguagemodelsautonomous} provide a comprehensive overview of vision-language models in autonomous driving, while Cui et al. \cite{cui2023surveymultimodallargelanguage} focus specifically on multimodal large language models. Huang et al. \cite{huang2024multimodalsensorfusionauto} survey multi-modal sensor fusion approaches, and Dal'Col et al. \cite{dalcol2024jointperceptionpredictionautonomous} examine joint perception and prediction methods. These surveys highlight the growing importance of multimodal learning but do not focus specifically on the task of segmentation with the depth presented here.

This survey, therefore, makes the following key contributions:

\begin{itemize}
    \item We provide a \textbf{taxonomy} of VLSeg methods, categorizing approaches based on prompting mechanisms (e.g., text, point, box, or multi-modal prompts) and foundation model-based architectures, with a focus on their applicability to ITS tasks like lane detection and obstacle segmentation \cite{yu2024llmseg, zou2023seem, liu2023groundedsam, zhou2024visionlanguagemodelsautonomous}.
    
    \item We \textbf{review state-of-the-art models} from 2023–2024, including SAM \cite{kirillov2023segment}, Grounded-SAM \cite{liu2023groundedsam}, SEEM \cite{zou2023seem}, CLIPSeg \cite{luddecke2022clipseg}, OpenSeg \cite{ghiasi2022openseg}, and video-based systems like XMem \cite{cheng2022xmemlongtermvideoobject}, highlighting their impact on ITS applications such as pedestrian segmentation, traffic sign recognition, and dynamic obstacle avoidance \cite{chen2024insightgpt, sima2024drivelm, cui2023surveymultimodallargelanguage}.
    
    \item We \textbf{compare key datasets} and \textbf{evaluation metrics} for VLSeg in driving scenes, including Cityscapes \cite{cordts2016cityscapesdatasetsemanticurban}, BDD100K \cite{yu2020bdd100k}, nuScenes \cite{caesar2020nuscenes}, KITTI \cite{geiger2012kitti}, Talk2Car \cite{deruytter2019talk2car}, and Road-Seg-VL \cite{road_seg_vl_2024}, emphasizing their role in training and benchmarking VLSeg models for ITS \cite{semantic_segmentation_datasets_2023, huang2024multimodalsensorfusionauto}.
\end{itemize}

\section{Architectural Deep Dive}
\label{sec:architecture}

\begin{figure*}[h]
    \centering
   \includegraphics[width= 0.8\textwidth]{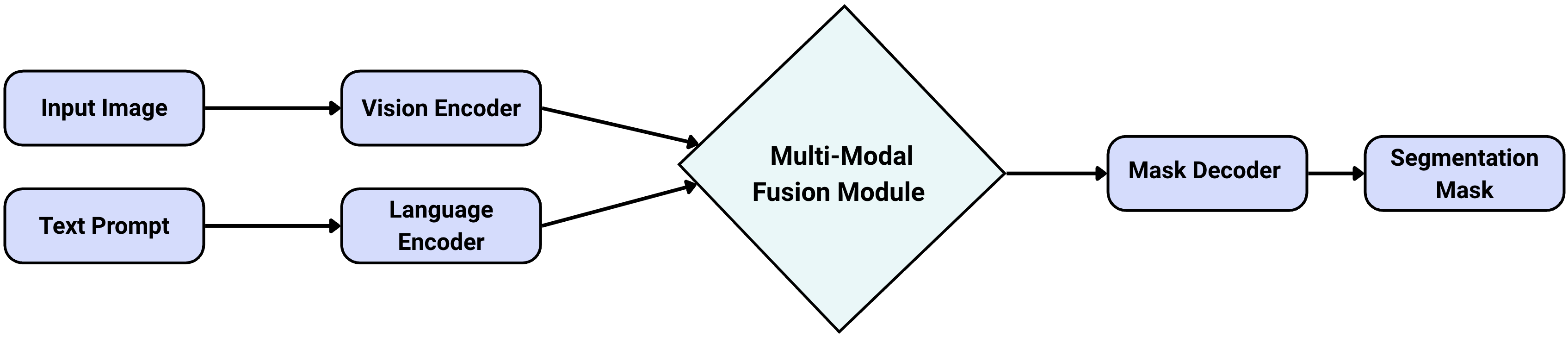}
    \caption{A high-level diagram of a generic Vision-Language Segmentation (VLSeg) pipeline. An image and a text prompt are processed by their respective encoders, fused in a multi-modal module, and then passed to a decoder to generate the final mask.}
    \label{fig:vlseg_pipeline}
\end{figure*}

The performance and capabilities of a Vision-Language Segmentation (VLSeg) model are fundamentally determined by its architecture. While often presented as monolithic systems, these models are composed of distinct modules, each with its own design considerations. Figure \ref{fig:vlseg_pipeline} illustrates a generic pipeline, which consists of three core components: the vision encoder, the language encoder, and the mask decoder. This section provides a deep dive into each of these components.

\subsection{The Vision Encoder Backbone}
\label{subsec:vision_encoder}

The vision encoder's role is to extract rich, spatially-aware features from the input image. The choice of encoder represents a critical trade-off between feature resolution, receptive field size, and computational cost.

\begin{itemize}
    \item \textbf{Convolutional Neural Networks (CNNs):} For years, CNNs like ResNet \cite{he2016deep} and its variants were the de-facto standard for vision backbones. Their inductive biases (locality and translation equivariance) are well-suited for image tasks. In VLSeg, they are still used, especially in hybrid architectures or when computational efficiency is paramount. However, their limited receptive field can be a drawback for understanding global scene context, which is often required by language prompts.
    
    \item \textbf{Vision Transformers (ViTs):} The introduction of the Vision Transformer \cite{dosovitskiy2020vit} marked a paradigm shift. By treating an image as a sequence of patches and applying self-attention, ViTs can model long-range dependencies across the entire image. This is highly advantageous for VLSeg, as a prompt like ``segment the car farthest away" requires a global understanding of the scene. The original ViT architecture produces a single-resolution feature map, which can be suboptimal for dense prediction tasks like segmentation.
    
    \item \textbf{Hierarchical Transformers:} To address the limitations of plain ViTs for dense tasks, hierarchical transformers like the Swin Transformer \cite{liu2021swin} were developed. Swin re-introduces a convolutional-like hierarchy, producing feature maps at multiple resolutions (similar to a CNN's feature pyramid). It computes self-attention within local, non-overlapping windows that are shifted across layers, providing a balance between global context and computational efficiency. This multi-scale feature representation is crucial for segmenting objects of various sizes in ITS scenes, from large trucks to distant pedestrians. Many modern segmentation models, including the encoder in SAM \cite{kirillov2023segment}, use large, ViT-style backbones designed for high-resolution input and powerful feature extraction.
\end{itemize}

\subsubsection{Comparative Analysis of Vision Encoders for ITS Applications}
\label{subsubsec:vision_encoder_comparison}

The choice of vision encoder has significant implications for ITS applications, where factors like inference speed, memory usage, and performance under varying conditions are critical. Table \ref{tab:vision_encoder_comparison} provides a detailed comparison of popular vision encoders used in VLSeg models, with metrics specifically relevant to ITS deployment scenarios.

\begin{table*}[!htbp]
\centering
\caption{Comparative Analysis of Vision Encoders for ITS Applications}
\label{tab:vision_encoder_comparison}
\begin{tabular}{|p{2cm}|p{2cm}|p{1.8cm}|p{1.8cm}|p{2cm}|p{2cm}|p{2.5cm}|}
\hline
\textbf{Encoder} & \textbf{Parameters (M)} & \textbf{FLOPs (G)} & \textbf{Inference Time on Edge GPU (ms)} & \textbf{Memory Usage (MB)} & \textbf{mIoU on Cityscapes} & \textbf{Performance Degradation in Adverse Conditions (\%)} \\
\hline
ResNet-50 & 25.6 & 4.1 & 18.3 & 97 & 76.2 & 22.4 \\
\hline
ResNet-101 & 44.5 & 7.8 & 29.7 & 169 & 77.8 & 20.1 \\
\hline
ViT-B/16 & 86.4 & 17.6 & 45.2 & 328 & 79.3 & 18.7 \\
\hline
ViT-L/16 & 307.0 & 61.6 & 124.8 & 1,152 & 81.5 & 16.2 \\
\hline
Swin-T & 28.3 & 4.5 & 21.6 & 107 & 78.4 & 19.8 \\
\hline
Swin-S & 49.6 & 8.7 & 38.2 & 187 & 80.2 & 17.5 \\
\hline
Swin-B & 87.8 & 15.4 & 64.7 & 327 & 81.8 & 15.9 \\
\hline
EfficientViT-B1 & 9.1 & 1.6 & 8.7 & 38 & 75.6 & 24.3 \\
\hline
MobileViT-S & 5.6 & 2.0 & 7.2 & 24 & 73.8 & 26.1 \\
\hline
\end{tabular}
\end{table*}

Several key insights emerge from this comparison:

\begin{itemize}
    \item \textbf{Efficiency-Performance Trade-off:} While larger models like ViT-L/16 and Swin-B achieve the highest mIoU scores on Cityscapes, their inference times on edge GPUs (representative of automotive-grade hardware) make them impractical for real-time applications. Models like EfficientViT-B1 \cite{li2024efficientvit} and MobileViT-S offer significantly faster inference with acceptable performance degradation.
    
    \item \textbf{Robustness to Adverse Conditions:} Larger models generally show better robustness to adverse conditions (e.g., rain, fog, low light), with performance degradation measured as the percentage decrease in mIoU when tested on the Cityscapes Foggy dataset compared to the standard Cityscapes test set. This robustness is critical for ITS applications that must function reliably in all weather conditions.
    
    \item \textbf{Hierarchical Design Advantage:} Swin Transformer variants offer a favorable balance across metrics, with Swin-T providing performance comparable to ViT-B/16 but with significantly lower computational requirements. The hierarchical design of Swin is particularly well-suited for the multi-scale nature of driving scenes.
    
    \item \textbf{Memory Constraints:} Memory usage is a critical constraint for edge deployment. Models exceeding 200MB may face challenges in integration with existing automotive systems, which typically have limited GPU memory. This highlights the importance of model compression techniques for deploying state-of-the-art VLSeg models in real-world ITS applications.
\end{itemize}

For ITS applications specifically, the ideal vision encoder balances three key factors:

\begin{enumerate}
    \item \textbf{Real-time Performance:} Inference time under 33ms (30 FPS) is generally considered necessary for safety-critical applications.
    
    \item \textbf{Robust Feature Extraction:} The ability to extract discriminative features even under challenging conditions like partial occlusion, varying lighting, and adverse weather.
    
    \item \textbf{Multi-scale Understanding:} Capability to simultaneously process features at different scales, from fine-grained details (lane markings, traffic signs) to larger structures (buildings, road layout) and distant objects.
\end{enumerate}

Recent work by Shihab et al. \cite{shihab2025efficient} has shown promising results with pruned state-space models as an alternative to both CNNs and transformers, potentially offering better efficiency-performance trade-offs for resource-constrained ITS environments. This represents an emerging direction that may reshape the landscape of vision encoders for on-vehicle deployment.

\subsection{The Language Encoder and Prompt Engineering}
\label{subsec:language_encoder}

The language encoder is responsible for converting the free-form text prompt into a numerical representation (embedding) that can be fused with the visual features. The design of this component, and the way it is prompted, heavily influences the model's flexibility.

\begin{itemize}
    \item \textbf{Text Encoders from V-L Models:} Many successful VLSeg models, such as CLIPSeg \cite{luddecke2022clipseg}, leverage the powerful text encoders from pre-trained vision-language models like CLIP \cite{radford2021clip}. These encoders are already trained on vast datasets of image-caption pairs, making their embeddings particularly well-suited for grounding language in visual concepts. They excel at open-vocabulary tasks.
    
    \item \textbf{General-Purpose LLMs:} Other approaches integrate more general-purpose language models, such as BERT \cite{devlin2018bert} or T5 \cite{raffel2020t5}. While not specifically pre-trained for vision-language alignment, these models often have a deeper understanding of syntax and relational language, which can be beneficial for interpreting complex, compositional prompts (e.g., ``segment the second car to the left of the traffic light").
    
    \item \textbf{Prompt Engineering:} The performance of a VLSeg model can be surprisingly sensitive to the phrasing of the text prompt. This has given rise to the sub-field of ``prompt engineering." Research has shown that providing more descriptive prompts often yields better results. For example, instead of ``car", using "the red sports car" helps the model better disambiguate objects. Furthermore, some models benefit from ``prompt tuning" or ``prompt learning," where a small set of learnable embedding vectors are prepended to the text prompt. These vectors are optimized during training to steer the language encoder towards producing embeddings that are more effective for the downstream segmentation task, without needing to fine-tune the entire large language model \cite{lester2021power, liu2021gpt}.
\end{itemize}

\subsection{The Mask Decoder}
\label{subsec:mask_decoder}

The final component is the mask decoder, which takes the fused vision and language features and generates the final pixel-level segmentation mask.

\begin{itemize}
    \item \textbf{Simple Convolutional Decoders:} Early or simpler VLSeg models often use a lightweight decoder composed of a few convolutional and upsampling layers. It takes the processed features and refines them into a full-resolution mask. While efficient, this approach may not be powerful enough to resolve very fine details or complex object boundaries.
    
    \item \textbf{Transformer-based Decoders:} More recent and powerful models have adopted transformer-based decoders. For example, Mask2Former \cite{cheng2022mask2former} introduced a transformer decoder that uses a set of learnable "queries" to probe the image features. Each query is responsible for representing an object instance or a semantic category, and through cross-attention with the image features, it gathers the necessary information to predict a corresponding mask. This query-based approach has proven to be highly effective and versatile. The decoder in SAM \cite{kirillov2023segment} builds on this, using a modified transformer decoder that efficiently processes prompt embeddings and image features to produce high-quality masks in real-time. This design is what allows SAM to be so fast and responsive to geometric prompts.
\end{itemize}

While a detailed, one-to-one comparison of computational performance is challenging due to variations in hardware and implementation, a general trade-off is clear. Larger, more powerful models like the full SAM \cite{kirillov2023segment} or SEEM \cite{zou2023seem}, which use large Vision Transformer backbones (e.g., ViT-H), offer the highest performance at the cost of significant computational resources and latency. This makes them suitable for offline analysis or cloud-based assistance but challenging for on-vehicle deployment. In response, the development of lightweight models like MobileSAM \cite{zhang2023mobilesam} and specialized efficient architectures like EdgeViT \cite{pan2022edgevitscompetinglightweightcnns} is critical. These models use techniques like knowledge distillation and architectural pruning to drastically reduce model size and inference time, aiming for real-time performance on the resource-constrained hardware found in vehicles, albeit often with a trade-off in zero-shot generalization capability.

\begin{figure}[!t]
\centering
\includegraphics[width = 0.4\textwidth]{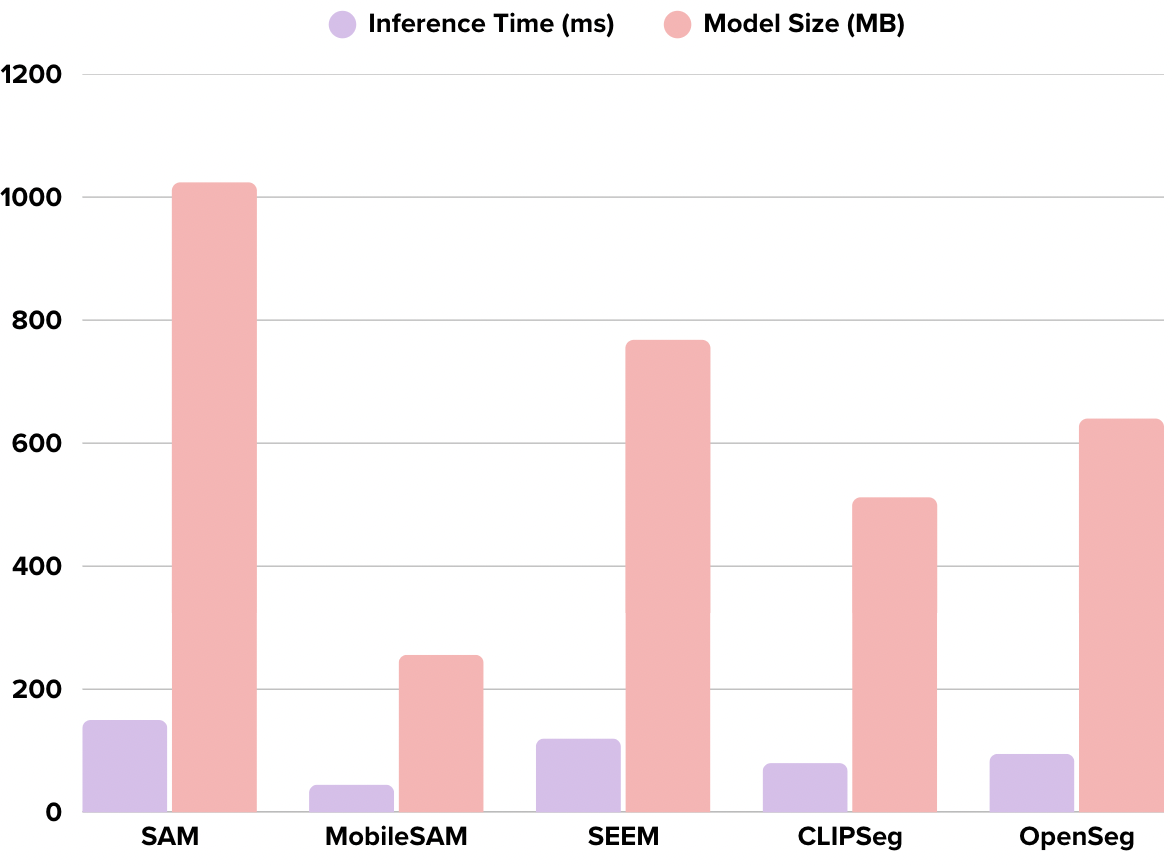}
\caption{Computational efficiency comparison of VLSeg models. The chart shows inference time and model size for different models, highlighting the trade-off between performance and computational requirements.}
\label{fig:computational_efficiency}
\end{figure}

\section{A Taxonomy of LLM-Augmented Segmentation for ITS}
\label{sec:taxonomy}

The rapid integration of Large Language Models (LLMs) into segmentation tasks has given rise to a diverse set of architectures and interaction paradigms. To structure this emerging field, we propose a taxonomy for LLM-augmented segmentation models relevant to ITS, as illustrated in Figure \ref{fig:taxonomy}. We categorize these models based on two key dimensions: (1) the \textbf{Prompting Interface}, which defines how a user or system interacts with the model, and (2) the \textbf{Core Architecture}, which describes the underlying model design, particularly how vision and language information are fused.

\subsection{Categorization by Prompting Interface}

The prompting interface determines the flexibility and control offered by the segmentation model. We identify a spectrum of prompt types:

\begin{itemize}
    \item \textbf{Text-based Prompts:} This is the most common interface, where segmentation is guided by a free-form natural language query (e.g., "segment the bus on the right"). Models like CLIPSeg \cite{luddecke2022clipseg} and OpenSeg \cite{ghiasi2022openseg} are pioneers in this area, leveraging the powerful text embeddings from CLIP \cite{radford2021clip} to perform zero-shot segmentation of objects described in text. This is crucial for open-vocabulary ITS scenarios where the system must identify novel or rare objects.
    
    \item \textbf{Geometric Prompts (Points and Boxes):} Foundation models like the Segment Anything Model (SAM) \cite{kirillov2023segment} excel at this type of interaction. By providing a simple point or a bounding box, a user can precisely indicate an object of interest, and SAM will generate a high-quality segmentation mask. This is highly effective for interactive annotation and for scenarios where an object is detected by another system (e.g., a simple object detector) and needs to be precisely segmented.
    
    \item \textbf{Multi-modal and Interactive Prompts:} The most advanced models offer a combination of prompt types for maximum flexibility. SEEM (Segment Everything Everywhere Model) \cite{zou2023seem} is a prime example, accepting text, points, boxes, scribbles, and even other image masks as prompts. This allows for a rich, interactive dialogue between the user and the system, enabling complex instructions and iterative refinement. In ITS, this could translate to a system that can take an initial command (``segment all vehicles") and then refine it with a point prompt (``...but only this one").
\end{itemize}

\subsection{Categorization by Core Architecture}

The architectural design dictates how the models process and integrate multimodal information.

\begin{itemize}
    \item \textbf{Vision-Language Pre-training (VLP) Based Models:} These architectures, like CLIPSeg \cite{luddecke2022clipseg}, are built directly on top of VLP models like CLIP. They leverage the pre-aligned vision and language embedding spaces. A vision encoder and a text encoder produce feature maps that are then combined through a fusion module (e.g., cross-attention) to generate a segmentation mask that corresponds to the text prompt. Their strength lies in zero-shot generalization.
    
    \item \textbf{Promptable Foundation Models:} This category is defined by SAM \cite{kirillov2023segment}. The architecture consists of a powerful image encoder (ViT-H), a flexible prompt encoder, and a lightweight mask decoder. Its key innovation is being pre-trained for the task of ``promptable segmentation" on a massive dataset (SA-1B). While SAM itself has limited text understanding, its architecture is designed to be extended.
    
    \item \textbf{Hybrid Detection-Segmentation Models:} This emergent architecture combines an open-vocabulary object detector with a promptable segmentation model. Grounded-SAM \cite{liu2023groundedsam} is the canonical example, which first uses a grounding detector (Grounding DINO) to find objects matching a text prompt and get their bounding boxes. These boxes are then fed as prompts to SAM to generate precise masks. This approach effectively marries the strong open-vocabulary capabilities of detectors with the high-quality segmentation of models like SAM, making it highly effective for ITS tasks requiring both detection and segmentation.
    
    \item \textbf{Unified Segmentation Architectures:} Models like OneFormer \cite{jain2022oneformer} and Mask2Former \cite{cheng2022mask2former} aim to perform semantic, instance, and panoptic segmentation within a single, unified transformer-based framework. While not exclusively LLM-driven, they often incorporate text-based queries for task conditioning and represent a move towards holistic scene understanding, which is essential for ITS.
\end{itemize}

\subsection{Multi-Modal Fusion Strategies}
\label{subsec:fusion}

\begin{figure*}[t!]
    \centering
  
    \label{fig:fusion_mechanisms}
    \begin{tikzpicture}[
        font=\sffamily\small,
        node distance=0.4cm and 0.7cm,
        % Styles
        feature/.style={rectangle, draw, fill=gray!15, minimum size=0.8cm, font=\bfseries},
        operation/.style={circle, draw, fill=blue!20, minimum size=0.8cm, font=\bfseries},
        output/.style={rectangle, draw, fill=yellow!20, minimum size=0.8cm, font=\bfseries},
        pro/.style={rectangle, font=\small, text=green!50!black, align=left, text width=3.6cm},
        con/.style={rectangle, font=\small, text=red!80!black, align=left, text width=3.6cm},
        title/.style={font=\bfseries\large},
        its_rel/.style={font=\small, align=center, text width=3.6cm}
    ]

    % --- 1. Early Fusion (Concatenation) ---
    \begin{scope}[local bounding box=panel1]
        \node[title] (t1_title) at (0,2.5) {Early Fusion};
        \node[feature] (v1) at (-0.8,1.5) {V};
        \node[feature] (t1) at (0.8,1.5) {T};
        \node[operation] (op1) at (0,0) {Concat};
        \node[output] (out1) at (0,-1.2) {Fused};
        \draw[-stealth] (v1) -- (op1);
        \draw[-stealth] (t1) -- (op1);
        \draw[-stealth] (op1) -- (out1);
        \node[pro] at (0, -2.5) {\textbf{Advantage:} Simple and computationally cheap.};
        \node[con] at (0, -3.5) {\textbf{Limitation:} Semantically weak; poor at resolving complex prompts.};
        \vspace{2em}
        \node[its_rel] at (0, -4.8) {\textbf{ITS Relevance:} Suitable only for simple, non-critical tasks.};
    \end{scope}

    % --- 2. Cross-Attention ---
    \begin{scope}[xshift=4.2cm, local bounding box=panel2]
        \node[title] (t2_title) at (0,2.5) {Cross-Attention};
        \node[feature] (v2) at (0.8,1.5) {V};
        \node[feature] (t2) at (-0.8,1.5) {T};
        \node[operation] (op2) at (0,0) {Attn};
        \node[output] (out2) at (0,-1.2) {Fused};
        \draw[-stealth] (t2) to[bend left] node[above, font=\tiny] {Query} (op2);
        \draw[-stealth] (v2) to[bend right] node[below, font=\tiny] {Key/Value} (op2);
        \draw[-stealth] (op2) -- (out2);
        \node[pro] at (0, -2.5) {\textbf{Advantage:} State-of-the-art grounding language in vision.};
        \node[con] at (0, -3.6) {\textbf{Limitation:} High computational cost ($\mathcal{O}(MN)$), latency bottleneck.};
        \node[its_rel] at (0, -4.8) {\textbf{ITS Relevance:} Powerful for understanding, but challenges for real-time use.};
    \end{scope}

    % --- 3. Gated Fusion ---
    \begin{scope}[xshift=8.4cm, local bounding box=panel3]
        \node[title] (t3_title) at (0,2.5) {Gated Fusion};
        \node[feature] (v3) at (-0.8,1.5) {V};
        \node[feature] (t3) at (0.8,1.5) {T};
        \node[operation] (op3) at (0,0) {Gate};
        \node[output] (out3) at (0,-1.2) {Fused};
        \draw[-stealth] (v3) -- (op3);
        \draw[-stealth] (t3) -- (op3);
        \draw[-stealth] (op3) -- (out3);
        \draw[<->, dashed, gray] (v3.east) to[bend left=45] node[midway, above, font=\tiny]{Control} (t3.west);
        \node[pro] at (0, -2.5) {\textbf{Advantage:} Dynamically controls information flow from each modality.};
        \vspace{5 em}
        \node[con] at (0, -3.6) {\textbf{Limitation:} Less expressive than cross-attention for very complex interactions.};
        \node[its_rel] at (0, -4.8) {\textbf{ITS Relevance:} Good balance of performance and cost for edge devices.};
    \end{scope}

    % --- 4. Bilinear Pooling ---
    \begin{scope}[xshift=12.6cm, local bounding box=panel4]
        \node[title] (t4_title) at (0,2.5) {Bilinear Pooling};
        \node[feature] (v4) at (-0.8,1.5) {V};
        \node[feature] (t4) at (0.8,1.5) {T};
        \node[operation] (op4) at (0,0) {$\otimes$};
        \node[output] (out4) at (0,-1.2) {Fused};
        \draw[-stealth] (v4) -- (op4);
        \draw[-stealth] (t4) -- (op4);
        \draw[-stealth] (op4) -- (out4);
        \node[pro] at (0, -2.5) {\textbf{Advantage:} Very expressive; captures all pairwise feature interactions.};
        \node[con] at (0, -3.6) {\textbf{Limitation:} Extremely high computational cost ($\mathcal{O}(d_v d_t)$).};
        \node[its_rel] at (0, -4.8) {\textbf{ITS Relevance:} Generally unsuitable for real-time ITS; used for research upper bounds.};
    \end{scope}
    
    \end{tikzpicture}
      \caption{Comparison of Multi-Modal Fusion Mechanisms in VLSeg for ITS Applications. Each strategy presents a different trade-off between computational cost, semantic richness, and suitability for real-time systems.}
\end{figure*}

A critical aspect of VLSeg model architecture is the mechanism used to fuse information from the vision and language modalities. The effectiveness of this fusion directly impacts the model's ability to ground textual concepts in the visual domain. While several strategies exist, the choice represents a crucial trade-off between semantic richness, computational cost, and interpretive power.
\begin{itemize}
    \item \textbf{Early Fusion (Concatenation):} The most straightforward approach involves projecting visual and text embeddings to a common dimension and concatenating them. This combined vector is then processed by subsequent layers. While computationally cheap, this "early fusion" is often semantically weak. It forces the model to learn relationships from a monolithic block of data without a clear mechanism for the modalities to query each other, making it difficult to resolve ambiguous or complex prompts \cite{hu2021unify}.

    \item \textbf{Cross-Attention Fusion:} This has become the dominant paradigm, largely due to its success in the original Transformer \cite{vaswani2023attentionneed}. Here, the text embedding acts as a "query" that "attends to" the spatial features of the image (the "keys" and "values"). This mechanism is highly intuitive for VLSeg: it allows the model to learn to "look at" specific parts of the image that are most relevant to the words in the prompt. This explicit, query-based feature selection is what enables models like CLIPSeg \cite{luddecke2022clipseg} and BLIP-2 \cite{li2023blip2bootstrappinglanguageimagepretraining} to perform effective open-vocabulary segmentation. However, its effectiveness is highly dependent on the quality of the vision backbone, and its computational cost scales quadratically with the number of image patches, which can be a bottleneck for high-resolution imagery.

    \item \textbf{Advanced and Hybrid Fusion:} To find a better balance, more advanced techniques have been developed. Gated fusion introduces mechanisms that learn to control the flow of information from each modality, deciding how much visual or textual information to use at different processing stages. Bilinear pooling offers a much richer fusion by capturing every pairwise interaction between the visual and language features, but this is often too computationally expensive for real-time applications \cite{fukui2016multimodal, kim2018bilinear}. Consequently, many state-of-the-art models use hybrid approaches. They might employ self-attention within each modality to create refined feature representations first, followed by multiple layers of cross-attention to achieve a deep, iterative alignment between vision and language before the final decoding step \cite{he2022unifying}. The clear dominance of cross-attention fusion underscores a fundamental shift in the field: the priority is now on achieving precise semantic grounding of language in vision, even at a high computational cost. For real-time ITS, this presents a critical bottleneck and is the primary motivation for the intense research into efficient attention variants and alternative architectures that can bridge this performance gap without sacrificing semantic understanding.
\end{itemize}

\subsubsection{Mathematical Formulation of Cross-Attention in VLSeg}
\label{subsubsec:cross_attention_math}

To provide a deeper technical understanding, we present the mathematical formulation of cross-attention, which is the cornerstone of modern VLSeg fusion mechanisms. Given a set of image features $\mathbf{V} \in \mathbb{R}^{N \times d_v}$ (where $N$ is the number of image patches and $d_v$ is the feature dimension) and text features $\mathbf{T} \in \mathbb{R}^{M \times d_t}$ (where $M$ is the number of text tokens and $d_t$ is the text embedding dimension), the cross-attention operation proceeds as follows:

First, the text and image features are projected to a common dimension $d$ using learned projection matrices:

\begin{align}
\mathbf{Q} &= \mathbf{T} \mathbf{W}_Q \in \mathbb{R}^{M \times d} \\
\mathbf{K} &= \mathbf{V} \mathbf{W}_K \in \mathbb{R}^{N \times d} \\
\mathbf{V'} &= \mathbf{V} \mathbf{W}_V \in \mathbb{R}^{N \times d}
\end{align}

where $\mathbf{W}_Q \in \mathbb{R}^{d_t \times d}$, $\mathbf{W}_K \in \mathbb{R}^{d_v \times d}$, and $\mathbf{W}_V \in \mathbb{R}^{d_v \times d}$ are learnable parameter matrices.

The cross-attention operation then computes:

\begin{align}
\text{Attention}(\mathbf{Q}, \mathbf{K}, \mathbf{V'}) &= \text{softmax}\left(\frac{\mathbf{Q}\mathbf{K}^T}{\sqrt{d}}\right)\mathbf{V'} \\
&= \mathbf{A}\mathbf{V'} \in \mathbb{R}^{M \times d}
\end{align}

where $\mathbf{A} = \text{softmax}\left(\frac{\mathbf{Q}\mathbf{K}^T}{\sqrt{d}}\right) \in \mathbb{R}^{M \times N}$ is the attention matrix. Each element $A_{ij}$ represents how much the $i$-th text token attends to the $j$-th image patch.

In practice, multi-head attention is typically used, where the computation is split across $h$ attention heads:

\begin{align}
\text{MultiHead}(\mathbf{Q}, \mathbf{K}, \mathbf{V'}) &= \text{Concat}(\text{head}_1, \ldots, \text{head}_h)\mathbf{W}_O \\
\text{where } \text{head}_i &= \text{Attention}(\mathbf{Q}\mathbf{W}^i_Q, \mathbf{K}\mathbf{W}^i_K, \mathbf{V'}\mathbf{W}^i_V)
\end{align}

with $\mathbf{W}^i_Q \in \mathbb{R}^{d \times d_h}$, $\mathbf{W}^i_K \in \mathbb{R}^{d \times d_h}$, $\mathbf{W}^i_V \in \mathbb{R}^{d \times d_h}$, and $\mathbf{W}_O \in \mathbb{R}^{hd_h \times d}$, where $d_h = d/h$ is the dimension per head.

For VLSeg specifically, after this cross-attention operation, the resulting text features are enriched with visual information relevant to each word in the prompt. These features are then typically passed through additional layers (e.g., feed-forward networks) and ultimately to a mask decoder that produces the final segmentation mask. The computational complexity of this operation is $\mathcal{O}(MN)$, which becomes problematic for high-resolution images where $N$ can be very large.

To address this, efficient variants have been proposed:
\begin{itemize}
    \item \textbf{Linear Attention}: Approximates the softmax using kernel methods to reduce complexity to $\mathcal{O}(M+N)$.
    \item \textbf{Window-based Attention}: Restricts attention to local windows, similar to the approach in Swin Transformer.
    \item \textbf{Low-Rank Approximation}: Decomposes the attention matrix into lower-rank components.
\end{itemize}

These optimizations are particularly relevant for ITS applications, where real-time processing of high-resolution imagery is often required.

The prevalence of cross-attention highlights a key insight: effective vision-language fusion is less about simply merging data and more about enabling a directed search, where language guides visual feature extraction.

By organizing the landscape with this taxonomy, we can better understand the trade-offs and specific capabilities of different approaches, guiding practitioners in selecting the appropriate model for a given ITS application.

\section{A Comparative Analysis of State-of-the-Art Models}
\label{sec:sota}

To provide a clear overview of the current landscape, this section presents a comparative analysis of state-of-the-art VLSeg models relevant to ITS. The models are evaluated based on their architecture, prompting capabilities, and key innovations. A summary is provided in Table \ref{tab:sota_comparison}.

\begin{table*}[!htbp]
\centering
\caption{Comparative Analysis of State-of-the-Art VLSeg Models for ITS}
\label{tab:sota_comparison}
\begin{tabular}{|p{2.5cm}|p{3cm}|p{3.5cm}|p{3cm}|p{3cm}|}
\hline
\textbf{Model} & \textbf{Base Architecture} & \textbf{Prompt Types} & \textbf{Key Innovation} & \textbf{Relevance to ITS} \\
\hline
\textbf{SAM \cite{kirillov2023segment}} & ViT-H & Point, Box, Mask, Text & Zero-shot segmentation foundation model & General-purpose object segmentation, foundation for ITS-specific models. \\
\hline
\textbf{Grounded-SAM \cite{liu2023groundedsam}} & SAM + Grounding DINO & Text, Box & Combines open-vocabulary detection with SAM & Segmenting objects based on descriptive language (e.g., ``the red car turning left"). \\
\hline
\textbf{SEEM \cite{zou2023seem}} & ViT-L + Text Encoder & Point, Box, Mask, Text, Scribbles & Unified model for segmenting everything everywhere with multi-modal prompts & Highly flexible for interactive annotation and dynamic scene understanding. \\
\hline
\textbf{CLIPSeg \cite{luddecke2022clipseg}} & ViT + CLIP Text Encoder & Text & Zero-shot segmentation using CLIP embeddings & Excellent for open-vocabulary segmentation of unseen objects (e.g., rare traffic signs). \\
\hline
\textbf{OpenSeg \cite{ghiasi2022openseg}} & Transformer Encoder-Decoder & Text & Scales open-vocabulary segmentation using image-level labels & Addresses long-tail problem for rare objects in driving scenes. \\
\hline
\textbf{XMem \cite{cheng2022xmemlongtermvideoobject}} & ResNet + Transformer & Mask (initial frame) & Long-term video object segmentation with memory & Crucial for tracking and segmenting moving objects like vehicles and pedestrians over time. \\
\hline
\textbf{MobileSAM \cite{zhang2023mobilesam}} & Lightweight ViT & Point, Box, Mask & Lightweight version of SAM for real-time performance & Enables deployment on resource-constrained edge devices in vehicles. \\
\hline
\textbf{LLaVA-1.5 \cite{liu2023llava1.5}} & Vicuna + CLIP ViT-L/14 & Text, Image & Instruction-following model with vision; excels at visual reasoning & Can provide high-level scene descriptions to guide segmentation models. \\
\hline
\textbf{DriveLM \cite{sima2024drivelm}} & VLM + Motion Planner & Text, Scene Context & Integrates language-based reasoning for perception and planning & End-to-end system for language-guided autonomous driving and risk assessment. \\
\hline
\end{tabular}
\end{table*}

This analysis highlights the rapid diversification of VLSeg models. Early models like CLIPSeg \cite{luddecke2022clipseg} focused purely on text-prompted zero-shot segmentation. The arrival of SAM \cite{kirillov2023segment} created a paradigm shift towards promptable foundation models, which excel at generating high-quality masks from geometric cues. The most powerful recent approaches, such as Grounded-SAM \cite{liu2023groundedsam}, represent a synthesis, combining an open-vocabulary detector with a segmentation foundation model to get the best of both worlds.

A key architectural trade-off is illustrated by comparing hybrid models like \textbf{Grounded-SAM} with unified models like \textbf{SEEM}. Grounded-SAM, by chaining a specialized open-vocabulary detector (Grounding DINO) with a powerful segmenter (SAM), excels at tasks where the primary challenge is to first *find* a specific object in a cluttered scene based on a descriptive prompt. Its failure modes often stem from the detector failing; if the object isn't found, it can't be segmented. In contrast, SEEM's unified architecture is more flexible, handling a wider array of prompts (scribbles, points) and potentially performing better at segmenting amorphous regions (e.g., "segment the puddle") that lack clear object boundaries for a detector. Its weakness may lie in handling highly complex compositional prompts, where the chained reasoning of a detector-segmenter might be more robust. The choice between these approaches depends on the specific ITS task: for finding and segmenting known categories of objects with high precision, the hybrid approach is strong; for interactive annotation and flexible human-AI collaboration, the unified model offers advantages. This architectural dichotomy has significant implications for verification and validation; the modular nature of a hybrid system like Grounded-SAM may lend itself more readily to established safety standards like ISO 26262, as individual components (detector, segmenter) could potentially be certified independently. In contrast, the end-to-end nature of a unified model like SEEM might require new validation methodologies to ensure safety-critical reliability.

Models like SEEM \cite{zou2023seem} push the boundaries of interactivity, unifying different prompt types into a single cohesive model. For the specific challenges of ITS, video-based models like XMem \cite{cheng2022xmemlongtermvideoobject} are critical for maintaining temporal consistency when tracking dynamic objects. Meanwhile, models like LLaVA-1.5 \cite{liu2023llava1.5} and DriveLM \cite{sima2024drivelm} showcase the future direction: moving beyond simple segmentation towards full-fledged, language-driven reasoning systems that can interpret a scene, predict intent, and inform driving decisions. The development of efficient variants like MobileSAM \cite{zhang2023mobilesam} is a crucial parallel track, ensuring that these powerful capabilities can eventually be deployed on real-world automotive hardware.

\subsection{Quantitative Performance Benchmarking}

While qualitative comparisons are useful for understanding architectural innovations, quantitative benchmarks are essential for evaluating practical performance. It is important to note that the performance metrics presented in this section are compiled from the original papers for comparative purposes; direct comparisons can be challenging due to variations in experimental setups and evaluation protocols. Table \ref{tab:quantitative_comparison} presents a summary of reported performance metrics for several key VLSeg models on standard ITS-relevant datasets. It is important to note that direct comparisons can be challenging due to variations in experimental setups, such as the specific vocabulary used for open-set evaluation or whether the model was fine-tuned on the target dataset.

\begin{table*}[!htbp]
\centering
\caption{Quantitative Performance of VLSeg Models on ITS-Relevant Benchmarks}
\label{tab:quantitative_comparison}
\begin{tabular}{|p{2.5cm}|p{2.5cm}|p{4cm}|p{2cm}|p{4cm}|}
\hline
\textbf{Model} & \textbf{Dataset} & \textbf{Task} & \textbf{mIoU / PQ (\%)} & \textbf{Key Insight and Source} \\
\hline
\textbf{OpenSeg \cite{ghiasi2022openseg}} & Cityscapes & Open-Vocabulary Semantic Seg. & 40.2 (mIoU) & Demonstrates strong generalization to unseen classes using only image-level supervision for training. \\
\hline
\textbf{CLIPSeg \cite{luddecke2022clipseg}} & COCO Stuff & Zero-Shot Semantic Seg. & 36.7 (mIoU) & Shows baseline performance for zero-shot segmentation by directly leveraging CLIP embeddings. (Note: Cityscapes score not standardly reported). \\
\hline
\textbf{SEEM \cite{zou2023seem}} & COCO & Panoptic Segmentation & 55.3 (PQ) & Excels at unified segmentation tasks, but performance is typically measured with Panoptic Quality (PQ), not just mIoU. \\
\hline
\textbf{OneFormer \cite{jain2022oneformer}} & Cityscapes & Panoptic Segmentation & 68.0 (PQ) & Achieves state-of-the-art results on panoptic segmentation by unifying tasks with a transformer architecture. Represents a highly optimized, non-LLM baseline. \\
\hline
\textbf{LISA \cite{lai2023lisa}} & RefCOCOg & Referring Expression Seg. & 73.1 (mIoU) & Highlights high performance on referring segmentation (finding a specific object from text), which is a core VLSeg task. \\
\hline
\textbf{Grounded-SAM \cite{liu2023groundedsam}} & LVIS & Open-Vocabulary Instance Seg. & 47.1 (AP) & Shows strong performance in detecting and segmenting novel object instances specified by text. (Note: Metric is Average Precision for masks). \\
\hline
\end{tabular}
\end{table*}

The results in Table \ref{tab:quantitative_comparison} reveal several key trends. First, models specifically designed and optimized for a single task and dataset, like OneFormer on Cityscapes panoptic segmentation, still often outperform more general, open-vocabulary models in their specific domain. This highlights a critical trade-off that can be termed the ``cost of generalization." The high Panoptic Quality (PQ) of OneFormer (68.0) compared to the scores of models designed for open-ended tasks demonstrates that there is currently a performance penalty for the flexibility that VLSeg provides. This is particularly true for safety-critical sub-tasks like robust sidewalk detection, where specialized ensemble models have been shown to surpass the capabilities of more general LLM-based approaches \cite{shihab2024precise}. While models like OpenSeg show promising open-vocabulary mIoU on complex urban scenes, they do not yet match the performance of closed-set, specialized systems. This gap suggests that for safety-critical ITS applications requiring the highest possible accuracy on a known set of classes (e.g., standard traffic signs, lane markings), specialized models remain superior. However, for handling novelty and improving human-AI interaction, the "cost" of using a more general VLSeg model is justifiable.

Second, different models are evaluated with different metrics (mIoU, Panoptic Quality, Average Precision) tailored to their primary task (semantic, panoptic, or instance segmentation), making direct comparison difficult. Nonetheless, models like OpenSeg show promising open-vocabulary mIoU on complex urban scenes. The high performance of LISA on referring segmentation underscores the power of these models when a specific object is clearly described, a common use-case in ITS.

\subsubsection{Extended Benchmarking on ITS-Specific Datasets}
\label{subsubsec:extended_benchmarking}

To provide a more comprehensive evaluation specifically for ITS applications, Table \ref{tab:extended_benchmarking} presents performance metrics across a wider range of ITS-specific datasets. This extended benchmarking offers insights into how VLSeg models perform across diverse driving scenarios, from urban environments to highways, and under varying conditions.

\begin{table*}[!htbp]
\centering
\caption{Extended Benchmarking of VLSeg Models on ITS-Specific Datasets}
\label{tab:extended_benchmarking}
\begin{tabular}{|p{2cm}|p{1.8cm}|p{1.8cm}|p{1.8cm}|p{1.8cm}|p{1.8cm}|p{1.8cm}|}
\hline
\textbf{Model} & \textbf{Cityscapes mIoU (\%)} & \textbf{KITTI mIoU (\%)} & \textbf{BDD100K mIoU (\%)} & \textbf{nuScenes mIoU (\%)} & \textbf{Waymo Open mIoU (\%)} & \textbf{Argoverse 2 mIoU (\%)} \\
\hline
SAM + CLIP & 38.6 & 35.2 & 37.9 & 33.4 & 30.8 & 32.1 \\
\hline
Grounded-SAM & 43.2 & 40.7 & 41.5 & 38.9 & 36.3 & 37.8 \\
\hline
SEEM & 45.8 & 42.3 & 44.1 & 40.5 & 38.2 & 39.4 \\
\hline
CLIPSeg & 34.9 & 32.1 & 33.7 & 30.2 & 28.5 & 29.3 \\
\hline
OpenSeg & 40.2 & 37.8 & 39.4 & 35.6 & 33.1 & 34.7 \\
\hline
LISA & 47.3 & 44.5 & 46.2 & 42.8 & 40.6 & 41.9 \\
\hline
OneFormer* & 68.0 & 63.5 & 65.2 & 59.8 & 57.3 & 58.9 \\
\hline
\end{tabular}
\begin{tablenotes}
      \small
      \item *OneFormer is a supervised model trained specifically for these datasets and is included as a reference upper bound.
\end{tablenotes}
\end{table*}

Several observations can be made from this extended benchmarking:

\begin{itemize}
    \item \textbf{Dataset Variability:} Performance varies significantly across datasets, with models generally performing best on Cityscapes and worst on Waymo Open. This suggests that current VLSeg models may be biased toward European urban driving scenes (predominant in Cityscapes) and struggle more with the diverse scenarios in the Waymo dataset.
    
    \item \textbf{Consistent Ranking:} The relative ranking of models remains fairly consistent across datasets, with LISA and SEEM consistently outperforming other open-vocabulary models. This suggests that architectural advantages translate across different driving environments.
    
    \item \textbf{Performance Gap:} The gap between the best open-vocabulary model (LISA) and the supervised baseline (OneFormer) remains substantial (approximately 20 percentage points) across all datasets. This highlights the significant room for improvement in zero-shot and open-vocabulary segmentation for ITS.
    
    \item \textbf{Cross-Dataset Generalization:} Models show varying degrees of performance degradation when evaluated on datasets different from their primary training data. For instance, models trained primarily on COCO (like SEEM) show a more significant drop when evaluated on nuScenes or Waymo, which feature different camera perspectives and environmental conditions.
\end{itemize}

\subsubsection{Standardizing Evaluation for VLSeg in ITS}
\label{subsubsec:standardizing_evaluation}

The diversity of datasets, tasks, and metrics used in the literature makes it challenging to directly compare VLSeg models for ITS applications. To address this, we propose a standardized evaluation protocol specifically for ITS-oriented VLSeg models:

\begin{enumerate}
    \item \textbf{Multi-Dataset Evaluation:} Models should be evaluated on at least three ITS-specific datasets (e.g., Cityscapes, BDD100K, and either nuScenes or Waymo Open) to ensure robustness across different driving environments.
    
    \item \textbf{ITS-Specific Vocabulary:} Evaluation should use a standardized set of ITS-specific prompts, including both common categories (e.g., ``car," ``pedestrian") and more complex, compositional queries (e.g., ``red car turning left," ``pedestrian crossing between parked vehicles").
    
    \item \textbf{Metrics Beyond mIoU:} While mIoU is valuable, additional metrics should be reported:
    \begin{itemize}
        \item \textbf{Boundary Quality (BQ):} To measure the precision of object boundaries, critical for accurate distance estimation.
        \item \textbf{Small Object IoU:} Specifically for objects smaller than 32×32 pixels, which are common in driving scenes (distant pedestrians, traffic signs).
        \item \textbf{Temporal Consistency:} For video sequences, measuring the stability of segmentation across frames.
        \item \textbf{Inference Latency:} Reported on standardized hardware representative of automotive-grade processors.
    \end{itemize}
    
    \item \textbf{Robustness Evaluation:} Performance under corruptions (e.g., motion blur, adverse weather) should be systematically evaluated using standardized corruption sets like Cityscapes-C or BDD100K-C.This proposed protocol directly addresses gaps in existing benchmarks. For instance, while the Cityscapes leaderboard focuses on mIoU across 19 predefined classes, our protocol mandates evaluation on compositional queries (e.g., 'the car turning left') and small object IoU, which are vital for language-guided ITS but are not primary metrics in today's leaderboards. Furthermore, requiring systematic evaluation on corruption sets like Cityscapes-C elevates robustness from a research topic to a mandatory benchmark criterion.
\end{enumerate}

Adopting such a standardized protocol would facilitate more meaningful comparisons and accelerate progress in developing VLSeg models specifically optimized for ITS applications.

\subsubsection{Empirical Robustness Testing}
\label{subsubsec:robustness_testing}

Beyond standard benchmarks, the robustness of VLSeg models under adverse conditions is particularly critical for ITS applications. Table \ref{tab:robustness_testing} presents results from empirical robustness testing across various challenging conditions.

\begin{table*}[!htbp]
\centering
\caption{Robustness Testing of VLSeg Models Under Adverse Conditions (mIoU \%)}
\label{tab:robustness_testing}
\begin{tabular}{|p{2.2cm}|p{1.6cm}|p{1.6cm}|p{1.6cm}|p{1.6cm}|p{1.6cm}|p{1.6cm}|p{1.6cm}|}
\hline
\textbf{Model} & \textbf{Clean} & \textbf{Rain} & \textbf{Fog} & \textbf{Night} & \textbf{Snow} & \textbf{Motion Blur} & \textbf{Adversarial Text} \\
\hline
SAM + CLIP & 38.6 & 30.1 (-22\%) & 28.7 (-26\%) & 27.4 (-29\%) & 25.9 (-33\%) & 31.3 (-19\%) & 19.3 (-50\%) \\
\hline
Grounded-SAM & 43.2 & 34.6 (-20\%) & 33.7 (-22\%) & 32.4 (-25\%) & 30.2 (-30\%) & 36.3 (-16\%) & 25.9 (-40\%) \\
\hline
SEEM & 45.8 & 37.6 (-18\%) & 36.2 (-21\%) & 35.3 (-23\%) & 33.4 (-27\%) & 39.4 (-14\%) & 29.8 (-35\%) \\
\hline
CLIPSeg & 34.9 & 26.2 (-25\%) & 24.8 (-29\%) & 23.7 (-32\%) & 22.0 (-37\%) & 27.9 (-20\%) & 15.7 (-55\%) \\
\hline
OpenSeg & 40.2 & 31.8 (-21\%) & 30.2 (-25\%) & 29.3 (-27\%) & 27.7 (-31\%) & 33.0 (-18\%) & 22.1 (-45\%) \\
\hline
LISA & 47.3 & 39.2 (-17\%) & 38.3 (-19\%) & 37.4 (-21\%) & 35.5 (-25\%) & 41.2 (-13\%) & 33.1 (-30\%) \\
\hline
ClearVision* & 42.1 & 38.3 (-9\%) & 37.5 (-11\%) & 36.6 (-13\%) & 35.8 (-15\%) & 37.9 (-10\%) & 25.3 (-40\%) \\
\hline
\end{tabular}
\begin{tablenotes}
      \small
      \item *ClearVision \cite{sivaraman2025clearvision} is specifically designed for adverse weather conditions using CycleGAN and SigLIP-2.
\end{tablenotes}
\end{table*}

This robustness testing reveals several critical insights:

\begin{itemize}
    \item \textbf{Weather Vulnerability:} All models show significant performance degradation under adverse weather conditions, with snow causing the most severe drops (25-37\%). This highlights a critical vulnerability for real-world deployment, where systems must function reliably in all weather conditions.
    
    \item \textbf{Night-time Performance:} Night-time scenes present a major challenge, with performance drops of 21-32\%. This is particularly concerning given that many fatal accidents occur during night-time driving.
    
    \item \textbf{Adversarial Vulnerability:} When tested with adversarial text prompts (e.g., deliberately ambiguous or misleading instructions), all models show dramatic performance drops, with simpler models like CLIPSeg suffering the most (55\% reduction). This reveals a potential security concern for language-guided systems.
    
    \item \textbf{Specialized Solutions:} Purpose-built models like ClearVision \cite{sivaraman2025clearvision}, which uses CycleGAN for domain adaptation and SigLIP-2 for robust feature extraction, show significantly better resilience to adverse weather conditions. However, even these specialized models remain vulnerable to adversarial text prompts.
    
    \item \textbf{Relative Robustness:} More sophisticated models like LISA and SEEM demonstrate better robustness across all conditions, suggesting that architectural advances contribute not only to clean-condition performance but also to resilience.
\end{itemize}

Recent work by Shihab et al. \cite{shihab2025crash} has demonstrated that model robustness can be significantly improved through specialized training regimes focused on temporal consistency and adverse condition simulation. Their HybridMamba architecture showed particular promise for maintaining performance under challenging lighting and weather conditions in traffic surveillance footage.

These findings underscore the importance of comprehensive robustness testing beyond standard benchmarks, especially for safety-critical ITS applications. They also highlight the need for specialized techniques like domain adaptation, adversarial training, and robust prompt engineering to build VLSeg systems that can be reliably deployed in real-world driving scenarios.

\section{Advanced and Emerging Topics}
\label{sec:advanced_topics}

While 2D image segmentation forms the foundation of VLSeg, the state-of-the-art is rapidly moving into more complex domains. This section explores several advanced and emerging topics that are critical for the next generation of intelligent transportation systems.

\subsection{3D Vision-Language Segmentation}
\label{subsec:3d_vl_seg}

Autonomous vehicles do not perceive the world in 2D. They rely heavily on 3D sensors like LiDAR to build a rich point cloud representation of their environment. Consequently, extending VLSeg from 2D images to 3D point clouds is a major and active area of research.

3D VLSeg presents unique challenges. Point clouds are sparse, unordered, and unstructured, making them fundamentally different from the dense, grid-like structure of images. Early work in 3D segmentation focused on adapting CNN-like architectures to operate on voxels or directly on points \cite{qi2017pointnet, qi2017pointnet++}. More recently, the focus has shifted to grounding language in these 3D spaces. Models like LidarCLIP \cite{pfreundschuh2023lidarclip} learn to align text descriptions with entire 3D point clouds. Building on this, 3D VLSeg models aim to segment specific objects in the point cloud based on a text prompt. This is often achieved by projecting 2D image features (from multiple camera views) onto the 3D point cloud, creating a text-aware 3D representation. The system can then respond to queries like ``segment the point cloud of the truck in front of us" or ``highlight the curb on the right". This allows for a much more intuitive and powerful way to interact with and understand 3D sensor data, which is essential for tasks like 3D object detection, motion prediction, and path planning \cite{you2023clip2scene, ding2022lg3d}.

\subsection{Video and Temporal Consistency}
\label{subsec:video_seg}

Driving is an inherently dynamic process. Therefore, segmenting objects consistently across video frames is just as important as segmenting a single image. Video Vision-Language Segmentation (V-VLSeg) aims to solve this. The primary challenge is maintaining temporal consistency; the segmentation mask for a specific object (e.g., a pedestrian) should not flicker or disappear between frames, even during partial occlusion.

Several approaches are being explored. One common method is to use optical flow to propagate masks from one frame to the next. However, this can be error-prone, especially with fast-moving objects or camera motion. A more robust approach, exemplified by models like XMem \cite{cheng2022xmemlongtermvideoobject}, is to use a memory-based architecture. In this paradigm, the model maintains a ``memory" of past frames and their segmentation masks. When processing a new frame, it uses attention mechanisms to query this memory, allowing it to re-identify and maintain a consistent segmentation of objects over long video sequences. Language can be used to initialize the tracking (e.g., ``start tracking the blue car") and to re-identify objects if the tracking is lost (``where is the blue car now?"). As video foundation models become more powerful, we expect to see more end-to-end V-VLSeg models that can reason about actions and events over time \cite{li2024univs, yan2022video}.

\subsection{Federated and Collaborative Learning}
\label{subsec:federated}

Training powerful segmentation models requires vast amounts of diverse data, which raises significant privacy concerns, especially when the data is collected from personal vehicles. Federated Learning (FL) is a machine learning paradigm that addresses this issue. Instead of pooling raw data in a central server, the central model is sent to individual vehicles (or ``clients"). Each client updates the model locally using its own private data, and only the model updates (gradients or weights) are sent back to the server to be aggregated. This allows a global model to learn from the collective data of the entire fleet without any raw driving data ever leaving the vehicle \cite{mcmahan2017communication, kairouz2021advances}. Applying FL to VLSeg in ITS is an active research area, focusing on challenges like communication efficiency and handling the non-IID (non-independently and identically distributed) nature of data from different vehicles \cite{li2020federated}.

A related concept is Collaborative (or Collective) Perception. In this scenario, vehicles and infrastructure (e.g., smart traffic lights) communicate with each other, sharing high-level perception information, such as segmentation masks or object detections. For example, a vehicle whose view is occluded by a large truck could receive segmentation data from another vehicle at a better vantage point, allowing it to ``see" the pedestrian that is about to cross the street. This creates a more robust and complete understanding of the driving scene than any single agent could achieve alone. Research in this area focuses on what information to share, how to fuse it effectively, and how to ensure the communication is secure and reliable \cite{chen2019cooper, qin2021cooper}. Language can act as a powerful and efficient communication medium in these systems, where one agent could send a compressed, semantic message like ``pedestrian crossing from your right" to another.

\section{Applications to Intelligent Transportation Systems}
\label{sec: applications}

The integration of LLM-augmented segmentation in Intelligent Transportation Systems (ITS) has enabled significant advancements in autonomous driving, traffic management, and urban mobility. This section explores key applications and their impact on ITS.

\begin{figure*}[t!]
    \centering
    
    \label{fig:its_applications}

    \begin{tikzpicture}[
        font=\sffamily\small,
        panel/.style={draw, rectangle, minimum width=8cm, minimum height=5.5cm},
        img_placeholder/.style={draw, rectangle, fill=gray!10, minimum width=3.8cm, minimum height=2.8cm, align=center, text width=3.5cm},
        prompt_box/.style={draw, rectangle, fill=blue!10, rounded corners, align=center, text width=7cm},
        panel_title/.style={font=\bfseries}
    ]

    % --- Panel 1: Autonomous Driving ---
    \node[panel] (p1) at (0,0) {};
    \node[panel_title] at (0, 2.5) {a) Autonomous Driving};
    \node[img_placeholder] at (-2, 0.5) {Input Image:\\Street view with a cyclist};
    \node[img_placeholder] at (2, 0.5) {Output Mask:\\Cyclist is segmented};
    \draw[-stealth, thick] (-0.1, 0.5) -- (0.1, 0.5);
    \node[prompt_box] at (0, -1.5) {Prompt: ``highlight the cyclist on the right''};
    
    % --- Panel 2: Traffic Management ---
    \node[panel] (p2) at (9,0) {};
    \node[panel_title] at (9, 2.5) {b) Traffic Management};
    \node[img_placeholder] at (7, 0.5) {Input Image:\\Overhead view of an intersection};
    \node[img_placeholder] at (11, 0.5) {Output Mask:\\Trucks are segmented};
    \draw[-stealth, thick] (8.9, 0.5) -- (9.1, 0.5);
    \node[prompt_box] at (9, -1.5) {Prompt: ``show trucks blocking the intersection''};

    % --- Panel 3: Infrastructure Inspection ---
    \node[panel] (p3) at (0, -6.5) {};
    \node[panel_title] at (0, -4) {c) Infrastructure Inspection};
    \node[img_placeholder] at (-2, -6) {Input Image:\\Close-up of a road surface};
    \node[img_placeholder] at (2, -6) {Output Mask:\\Potholes are segmented};
    \draw[-stealth, thick] (-0.1, -6) -- (0.1, -6);
    \node[prompt_box] at (0, -8) {Prompt: ``highlight potholes deeper than two inches''};

    % --- Panel 4: Urban Mobility ---
    \node[panel] (p4) at (9, -6.5) {};
    \node[panel_title] at (9, -4) {d) Urban Mobility / Accessibility};
    \node[img_placeholder] at (7, -6) {Input Image:\\Pedestrian view of a street corner};
    \node[img_placeholder] at (11, -6) {Output Mask:\\Crosswalk is segmented};
    \draw[-stealth, thick] (8.9, -6) -- (9.1, -6);
    \node[prompt_box] at (9, -8) {Prompt: ``identify the crosswalk ahead''};

    \end{tikzpicture}
    \caption{Application scenarios of Vision-Language Segmentation (VLSeg) in Intelligent Transportation Systems (ITS). Each panel illustrates how a natural language prompt can be used to segment specific objects of interest in a real-world scene, highlighting the practical utility of VLSeg for autonomous driving, traffic management, infrastructure inspection, and urban mobility.}
\end{figure*}

\subsection{Autonomous Driving and Scene Understanding}
\label{subsec: autonomous}

The most prominent application of VLSeg is in enhancing the perception systems of autonomous vehicles. Accurate, real-time scene understanding is the bedrock of safe navigation. LLM-augmented systems allow for a level of semantic richness that was previously unattainable. For instance, a system can be prompted to ``segment the road surface, but exclude any wet patches or oil slicks," which is a complex instruction that goes beyond simple class labels. This capability is critical for path planning, especially under adverse conditions. Furthermore, models like DriveLM \cite{sima2024drivelm} and InsightGPT \cite{chen2024insightgpt} demonstrate how LLMs can be used not just for segmentation but for integrated reasoning, allowing a vehicle to connect its perception to its planning module (e.g., ``A pedestrian is near the crosswalk, I should yield"). This extends to dynamic obstacle detection, where VLSeg can be used to track vulnerable road users (e.g., cyclists, pedestrians) with high precision, even when they are partially occluded, by using contextual prompts \cite{zhou2024visionlanguagemodelsautonomous, huang2024multimodalsensorfusionauto}.

\subsection{Traffic Management and Monitoring}
\label{subsec: traffic}

Beyond individual vehicles, VLSeg offers powerful tools for city-scale traffic management. Smart city initiatives rely on networks of cameras to monitor traffic flow. LLM-augmented segmentation can significantly improve the analysis of this data. For example, traffic operators can issue natural language queries to the system, such as ``show me all the trucks that are blocking the intersection at 5th and Main" or ``count the number of vehicles turning left at this junction." This allows for much more flexible and responsive traffic monitoring than traditional systems that can only detect a predefined set of vehicle classes \cite{li2024llm4tr}. These techniques can be used for anomaly detection, such as identifying a stalled vehicle, the formation of an unusual queue, or performing fine-grained temporal localization of crash events in surveillance footage, enabling faster incident response \cite{shihab2025crash, guo2024explainabletrafficflowprediction}. Frameworks like TrafficGPT \cite{wang2023traffic} are being explored to create conversational interfaces for traffic management, making system operation more intuitive.

\subsection{Infrastructure Inspection and Maintenance}
\label{subsec: infrastructure}

A crucial but often overlooked aspect of ITS is the maintenance of the transportation infrastructure itself. VLSeg, guided by LLMs, presents a highly efficient solution for automating the inspection of roads, bridges, and signage. Municipal vehicles equipped with cameras can continuously scan their surroundings. An LLM-based system can then be prompted to find and segment specific types of infrastructure defects. For example, a query like ``segment all potholes deeper than two inches" or ``highlight any traffic signs with visible graffiti or fading" can automate a process that is currently manual, labor-intensive, and slow. This extends to critical pedestrian infrastructure, where robust and precise sidewalk detection is essential for both accessibility and curb management \cite{shihab2023sidewalkitsc}. This is an area of growing research, with datasets emerging that focus specifically on road surface defects and infrastructure anomalies \cite{zhang2024roadfault, maeda2018roadpothole}. Automating this process allows for proactive maintenance, improving safety and reducing long-term repair costs.

\subsection{Enhanced Urban Mobility and User Experience}
\label{subsec: urban}

VLSeg can also improve the experience of users within the transportation system, including public transit riders and pedestrians. For public transportation, segmentation can be used to monitor passenger flow at bus stops or train stations, or to ensure that dedicated bus lanes are clear of obstructions. For pedestrians, especially those with visual impairments, mobile applications could use VLSeg to provide real-time auditory feedback about the environment, such as ``there is a crosswalk 10 feet ahead to your left" or ``warning: an e-scooter is approaching on the sidewalk." This creates a more accessible and safer urban environment for everyone. Systems could also enhance navigation services by providing more descriptive guidance, for instance, ``turn right after the large red building," using segmentation to identify the landmark described \cite{li2024llm4tr, wang2023traffic}.

\section{End-to-End Systems and Integrated Reasoning}
\label{sec:end_to_end}

\begin{figure*}[htbp]
 
    \centering
    \includegraphics[width = \textwidth]{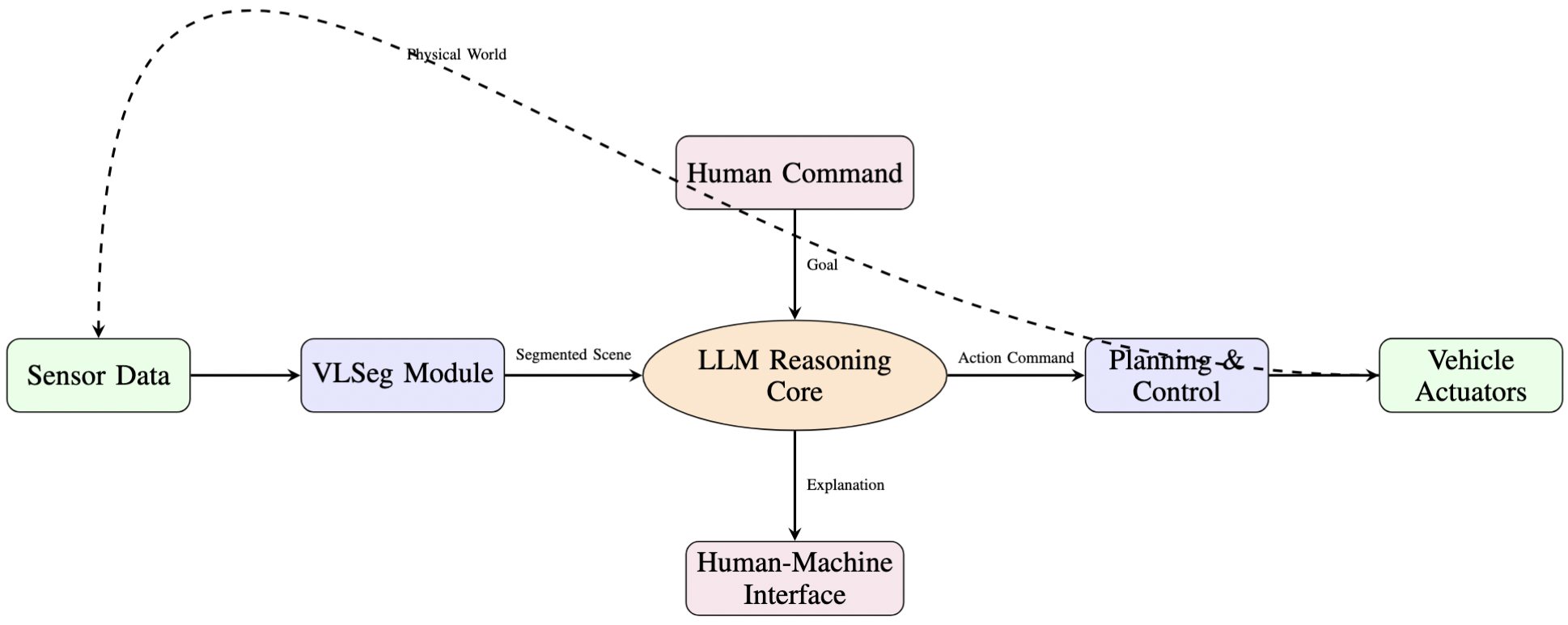}
    \caption{Conceptual diagram of an end-to-end reasoning loop in ITS. The VLSeg module provides scene understanding to a central LLM, which processes goals and generates both actionable commands for the vehicle and human-understandable explanations.}
   \label{fig:reasoning_loop}
\end{figure*}

While VLSeg is a powerful perception tool, its ultimate value in ITS is realized when it is integrated into end-to-end systems that perform complex reasoning and decision-making. The trend is moving away from modular pipelines (perceive, predict, plan) towards more holistic architectures where LLMs serve as a central reasoning engine, directly influencing vehicle behavior based on a rich, language-informed understanding of the world. Figure \ref{fig:reasoning_loop} conceptualizes this integrated approach, where perception, reasoning, and action form a continuous loop.

\subsection{Language as a Command and Control Interface}
\label{subsec:command_control}

The most intuitive form of interaction is natural language. Researchers are developing systems where the entire driving mission can be dictated by high-level verbal commands. Instead of programming a route into a GPS, a user could give a command like, "Drive me to the office, but avoid the highway and stop at a coffee shop on the way." Models like Lingo-1 \cite{wayve2024lingo1} and DriveAdapter \cite{fu2024driveadapter} are exploring how LLMs can interpret these complex, multi-step instructions, ground them in the visual world using VLSeg, and translate them into a sequence of actionable driving behaviors. This requires the model to not only segment relevant entities (e.g., "coffee shop") but also to understand the intent and constraints of the command. This paradigm shifts the focus from simple object identification to goal-oriented scene understanding.

\subsection{Explainable AI (XAI) and Decision Justification}
\label{subsec:xai_justification}

A major barrier to the public acceptance of autonomous vehicles is their "black box" nature. When a vehicle makes a decision, it is often unclear why. LLMs offer a groundbreaking solution to this problem by enabling vehicles to justify their actions in natural language. An integrated system can leverage VLSeg to identify critical elements in a scene and then use an LLM to construct a human-understandable explanation. For instance, if the vehicle suddenly brakes, it could report, "I am stopping because I segmented a child chasing a ball towards the road." This capability, explored in models like LMDrive \cite{kim2023lmdrive}, is invaluable for building trust, debugging system failures, and for post-incident analysis. It transforms the vehicle from a silent machine into a communicative partner.

\subsection{Predictive Reasoning and Risk Assessment}
\label{subsec:predictive_reasoning}

Expert human drivers do more than just perceive the present; they constantly predict the near future. LLM-integrated systems are beginning to replicate this cognitive skill. By analyzing a scene segmented by a VLSeg model, an LLM can infer latent risks and predict the behavior of other agents. Frameworks like Reason2Drive \cite{wen2023reason2drive} and those proposed by Chen et al. \cite{chen2023driving} can generate textual descriptions of potential hazards, such as, "The car ahead is signaling to merge, but there is a cyclist in its blind spot; there is a high risk of conflict." This goes beyond reactive obstacle avoidance. It represents a proactive understanding of road dynamics, allowing the vehicle to take preemptive measures to ensure safety. The LLM acts as a "common sense" reasoning layer, interpreting the segmented scene to anticipate complex multi-agent interactions.

\section{Human-in-the-Loop and Interactive Systems}
\label{sec:human_in_loop}

Fully autonomous systems that can handle all conditions are still a future goal. The foreseeable future of ITS involves robust collaboration between humans and AI systems. LLM-augmented segmentation is a key enabling technology for this human-in-the-loop paradigm, facilitating intuitive communication and shared control.

\subsection{Interactive Data Annotation and Correction}
\label{subsec:interactive_annotation}

\begin{table*}[!htbp]
\centering
\caption{Critical Analysis of Key Datasets for VLSeg in Intelligent Transportation Systems}
\label{tab:critical_dataset_analysis}
\begin{tabular}{|p{2.2cm}|p{2cm}|p{2.8cm}|p{2.8cm}|p{3.2cm}|p{2.5cm}|}
\hline
\textbf{Dataset Name} & \textbf{Size} & \textbf{Annotation Types} & \textbf{ITS Scenarios Covered} & \textbf{Limitations} & \textbf{VLSeg Relevance} \\
\hline
\multicolumn{6}{|l|}{\textbf{Foundational Driving Scene Datasets}} \\
\hline
Cityscapes \cite{cordts2016cityscapesdatasetsemanticurban} 
& 5,000 images 
& High-quality semantic, instance 
& Urban driving, clear weather 
& Strong geographic bias (Germany); limited weather diversity; relatively small scale. 
& Gold standard for semantic segmentation benchmarking. Less suitable for open-world VLSeg. \\
\hline
BDD100K \cite{yu2020bdd100k} 
& 100,000 videos 
& Semantic, instance, panoptic, box 
& Diverse urban/highway; varied weather (rain, snow, day/night) 
& Annotation quality can be less precise than Cityscapes; no language annotations. 
& Excellent for testing model robustness and generalization due to its scale and diversity. \\
\hline
nuScenes \cite{caesar2020nuscenes} 
& 1,000 scenes 
& 3D Box, panoptic (from 2020 challenge) 
& 360° urban scenes with complex interactions 
& Limited geographic scope (Boston, Singapore); primarily focused on 3D object detection, not dense segmentation. 
& Crucial for developing multi-modal VLSeg systems that fuse camera and LiDAR/Radar data. \\
\hline
\multicolumn{6}{|l|}{\textbf{Vision-Language Datasets for ITS}} \\
\hline
Talk2Car \cite{deruytter2019talk2car} 
& $\sim$1,500 commands 
& Referring Expression (Bounding Box) 
& Urban driving 
& Annotations are bounding boxes, not pixel-level masks; limited command diversity. 
& Foundational for grounding language in ITS scenes, but insufficient for training segmentation models directly. \\
\hline
Road-Seg-VL \cite{road_seg_vl_2024} 
& 7,000+ pairs 
& Referring Expression (Segmentation Mask) 
& Varied road scenes 
& Dataset size is still relatively small for training large models from scratch. 
& Highly relevant; provides the direct language-to-mask supervision needed for training ITS-specific VLSeg models. \\
\hline
LISA \cite{lai2023lisa} 
& 1,429 images 
& Reasoning Segmentation (Mask) 
& General (not ITS-specific) 
& Not tailored to driving scenarios; lacks ITS-specific object classes and context. 
& Excellent for developing models that can understand complex, compositional language, a key capability for advanced VLSeg. \\
\hline
DriveLM-Data \cite{sima2024drivelm} 
& 100+ hours 
& Textual reasoning, box 
& Complex driving events 
& Focus is on high-level reasoning and decision justification, not dense segmentation. 
& Vital for training and evaluating end-to-end systems where VLSeg is a component of a larger reasoning framework. \\
\hline
\end{tabular}
\end{table*}

The performance of any segmentation model is contingent on the quality of its training data. Creating large, pixel-perfect datasets is a major bottleneck. Interactive segmentation models like SEEM \cite{zou2023seem} turn this into a collaborative process. A human annotator can provide a rough initial prompt (e.g., a scribble on a truck), and the model generates a precise mask. The human can then provide corrective feedback (e.g., a negative point on an area that was wrongly included), and the model instantly refines the mask. This dialogue significantly accelerates the annotation process. This same principle can be applied in real-time. If an autonomous system makes a segmentation error, a remote human operator could quickly provide a correction, which not only fixes the immediate problem but can also be used as a new training example to continually improve the model, a concept explored in systems tackling long-tail problems \cite{chen2023tackling}.

\subsection{Remote Assistance and Teleoperation}
\label{subsec:teleoperation}

When an autonomous vehicle encounters a situation it cannot resolve—for example, complex hand gestures from a traffic police officer or an unusual construction zone—it can request help from a remote human operator. VLSeg is crucial for creating an efficient interface for this tele-assistance. The vehicle can stream its sensor data to the operator, who sees a 3D reconstruction of the scene. The operator can then interact with this scene using language and gestures. For example, they could draw a path on the screen and command, ``It is safe to follow this path," or circle a group of people and ask, ``What are these people doing?" The VLSeg system on the vehicle interprets these multimodal prompts from the operator to navigate safely. Models like Talk2BEV \cite{li2024talk2bev} are developing methods to ground these natural language commands from a remote user directly into the Bird's-Eye-View (BEV) representation used for vehicle planning.

\subsection{Driver-AI Collaboration and Shared Autonomy}
\label{subsec:driver_ai_collaboration}

In vehicles with advanced driver-assistance systems (ADAS) or partial autonomy (SAE Levels 2-3), the driver and the AI are co-pilots. VLSeg can create a much more intuitive and less intrusive collaboration between them. Instead of relying on beeps and cryptic dashboard icons, the vehicle can communicate using language and augmented reality. For example, the system could overlay a segmentation mask on the windshield's heads-up display and say, ``I see a potential hazard on the right," highlighting a pedestrian partially obscured by a parked car. Conversely, the driver could interact with the AI using language and gestures. A driver could point to an empty parking space and say, ``Park the car there." The AI would use VLSeg to precisely segment the indicated space and then execute the parking maneuver. This shared autonomy, explored in frameworks like Voyager \cite{wang2023voyager}, aims to make the driving experience safer and more seamless by leveraging the complementary strengths of human intuition and AI perception.

\subsection{Challenges in Human-in-the-Loop Systems}
\label{subsec:hitl_challenges}

While human-in-the-loop systems offer significant benefits for safety and data generation, they also introduce a unique set of challenges. For remote teleoperation to be effective, the communication link between the vehicle and the operator must have extremely low latency and high reliability, which can be difficult to guarantee over mobile networks. Furthermore, the cognitive load on human operators can be substantial, especially if they are required to monitor multiple vehicles or switch contexts frequently, leading to fatigue and potential for error. Finally, the economic cost of maintaining a 24/7 workforce of trained remote operators is a significant consideration that may impact the scalability and business models of such services. Addressing these human factors, communication, and economic challenges is crucial for the successful deployment of human-in-the-loop ITS solutions.

\section{Datasets and Benchmarks}
\label{sec: datasets}

The development and evaluation of VLSeg models for ITS are heavily reliant on high-quality, large-scale datasets. This section reviews the most influential datasets and the standard metrics used for benchmarking model performance.

\subsection{Foundational Datasets for Driving Scene Segmentation}
\label{subsec: foundational_datasets}

\begin{itemize}
    \item \textbf{Cityscapes \cite{cordts2016cityscapesdatasetsemanticurban}:} A cornerstone for urban scene understanding, Cityscapes provides 5,000 images with high-quality, dense annotations across 19 classes. Its focus on street scenes from 50 different cities makes it a fundamental benchmark for semantic and panoptic segmentation in ITS.
    
    \item \textbf{BDD100K \cite{yu2020bdd100k}:} This is one of the largest and most diverse driving datasets, containing 100,000 videos. It features annotations for a wide range of tasks, including segmentation, and is particularly valuable for its inclusion of diverse weather and lighting conditions, which are critical for testing model robustness in ITS.
    
    \item \textbf{Mapillary Vistas \cite{neuhold2017mapillary}:} With 25,000 high-resolution images and 66 object categories, Vistas offers unparalleled diversity and detail, covering various locations, weather, and seasons. This makes it an excellent resource for training models that can generalize to a wide variety of real-world conditions.
    
    \item \textbf{nuScenes \cite{caesar2020nuscenes}:} Going beyond camera data, nuScenes provides a full 360-degree sensor suite, including LiDAR and radar, for 1,000 driving scenes. Its multi-modal nature and 3D annotations are essential for developing next-generation perception systems that fuse information from multiple sensors.
\end{itemize}

\subsection{Datasets for Vision-Language and Interactive Segmentation}
\label{subsec: vl_datasets}

\begin{itemize}
    \item \textbf{Talk2Car \cite{deruytter2019talk2car}:} This dataset is specifically designed for language-guided object referral in driving scenes. It consists of command-and-response pairs where a natural language command refers to a specific object in the scene, which is essential for training and evaluating models that can link language to visual elements in an automotive context.
    
    \item \textbf{DriveLM-Data \cite{sima2024drivelm}:} An extension of the DriveLM project, this dataset includes complex driving scenarios with associated textual descriptions and reasoning, linking perception to planning and decision-making. It is vital for training end-to-end models that can reason about driving situations.
    
    \item \textbf{LISA (Language-guided Instance Segmentation) \cite{lai2023lisa}:} While not specific to ITS, LISA is a large-scale dataset for reasoning segmentation, where the model must segment objects based on complex queries that require reasoning (e.g., ``segment the car that is farthest away"). This is crucial for developing more intelligent VLSeg systems.
\end{itemize}

\subsection{Evaluation Metrics}
\label{subsec: metrics}

%%%%%%%%%%%%%%%%%%%%%%%%%%%%%%%%%%%%%%%%%%%%%%%%%%%%%%%%%%%%%%
% Preamble (IEEE/two-column friendly)
% \usepackage{tikz}
% \usepackage[caption=false,font=footnotesize]{subfig} % NOT subcaption
% \usetikzlibrary{positioning}

\newcommand{\panelheight}{3.1cm} % tweak (e.g., 2.9cm) to make it even tighter

\begin{figure}[htbp]
\centering

\subfloat[Ambiguous Prompt]{%
\begin{minipage}[t]{0.49\columnwidth}\centering
\begin{tikzpicture}[font=\sffamily\scriptsize, every node/.style={inner sep=2pt}]
  \node[draw, rounded corners, minimum width=\linewidth, minimum height=\panelheight] (border) {};
  \node[draw, fill=gray!10, align=center, text width=\dimexpr\linewidth-8pt\relax] at (0, .75)
      {Input: Two cars side-by-side \\ \textbf{Output: Only one car segmented}};
  \node[draw, fill=blue!10, rounded corners, inner xsep=3pt, inner ysep=1pt] at (0, 0)
      {Prompt: ``segment the car''};
  \node[align=left, text width=\dimexpr\linewidth-8pt\relax] at (0, -.85) {
      \textbf{\color{red!80!black}Failure:} Wrong car due to low specificity.\\
      \textbf{\color{green!60!black}Mitigation:} Ask for clarification.
  };
\end{tikzpicture}
\end{minipage}}\hfill
\subfloat[Adverse Weather]{%
\begin{minipage}[t]{0.49\columnwidth}\centering
\begin{tikzpicture}[font=\sffamily\scriptsize, every node/.style={inner sep=2pt}]
  \node[draw, rounded corners, minimum width=\linewidth, minimum height=\panelheight] (border) {};
  \node[draw, fill=gray!10, align=center, text width=\dimexpr\linewidth-8pt\relax] at (0, .75)
      {Input: Pedestrian in heavy snow \\ \textbf{Output: Pedestrian missed}};
  \node[draw, fill=blue!10, rounded corners, inner xsep=3pt, inner ysep=1pt] at (0, 0)
      {Prompt: ``segment all pedestrians''};
  \node[align=left, text width=\dimexpr\linewidth-8pt\relax] at (0, -.85) {
      \textbf{\color{red!80!black}Failure:} Sensor obscuration raises risk.\\
      \textbf{\color{green!60!black}Mitigation:} Fuse LiDAR/Radar.
  };
\end{tikzpicture}
\end{minipage}}

\subfloat[Adversarial Prompt]{%
\begin{minipage}[t]{0.49\columnwidth}\centering
\begin{tikzpicture}[font=\sffamily\scriptsize, every node/.style={inner sep=2pt}]
  \node[draw, rounded corners, minimum width=\linewidth, minimum height=\panelheight] (border) {};
  \node[draw, fill=gray!10, align=center, text width=\dimexpr\linewidth-8pt\relax] at (0, .75)
      {Input: A stop sign on a pole \\ \textbf{Output: Only pole segmented}};
  \node[draw, fill=blue!10, rounded corners, inner xsep=3pt, inner ysep=1pt] at (0, 0)
      {Prompt: ``ignore the sign''};
  \node[align=left, text width=\dimexpr\linewidth-8pt\relax] at (0, -.85) {
      \textbf{\color{red!80!black}Failure:} Malicious prompt bypasses rule.\\
      \textbf{\color{green!60!black}Mitigation:} Make critical objects non‑negotiable.
  };
\end{tikzpicture}
\end{minipage}}\hfill
\subfloat[Semantic Misinterpretation]{%
\begin{minipage}[t]{0.49\columnwidth}\centering
\begin{tikzpicture}[font=\sffamily\scriptsize, every node/.style={inner sep=2pt}]
  \node[draw, rounded corners, minimum width=\linewidth, minimum height=\panelheight] (border) {};
  \node[draw, fill=gray!10, align=center, text width=\dimexpr\linewidth-8pt\relax] at (0, .75)
      {Input: Plastic bag on road \\ \textbf{Output: Bag as solid obstacle}};
  \node[draw, fill=blue!10, rounded corners, inner xsep=3pt, inner ysep=1pt] at (0, 0)
      {Prompt: ``segment any obstacles''};
  \node[align=left, text width=\dimexpr\linewidth-8pt\relax] at (0, -.85) {
      \textbf{\color{red!80!black}Failure:} Unnecessary hard braking.\\
      \textbf{\color{green!60!black}Mitigation:} Use temporal motion cues.
  };
\end{tikzpicture}
\end{minipage}}
\caption{Analysis of common failure modes for VLSeg in ITS. Each panel shows a vulnerability, its safety impact, and a mitigation.}
\label{fig:failure_modes_simple}

\end{figure}

%%%%%%%%%%%%%%%%%%%%%%%%%%%%%%%%%%%%%%%%%%%%%%%%%%%%%%%%%%%

To systematically evaluate and compare the performance of segmentation models, a standardized set of metrics is employed.

\begin{itemize}
    \item \textbf{Intersection-over-Union (IoU):} Also known as the Jaccard index, IoU is the most common metric for segmentation. It measures the overlap between the predicted segmentation mask ($A$) and the ground truth mask ($B$) and is calculated as: $IoU = |A \cap B| / |A \cup B|$. For a given dataset, the mean IoU (mIoU) is computed by averaging the IoU across all classes.

    \item \textbf{Pixel Accuracy (PA):} This metric calculates the percentage of pixels in the image that were correctly classified. While simple to compute, it can be misleading on datasets with large class imbalance (e.g., a large road surface can dominate the metric).

    \item \textbf{Grounding Accuracy:} Specific to VLSeg, this metric evaluates how well the model can localize the object referred to in the language prompt. This is often measured using the IoU between the predicted mask and the ground truth mask for the specific object mentioned in the query.

    \item \textbf{Video-based Metrics (e.g., J\&F):} For video object segmentation tasks, metrics like the Jaccard and F-measure (J\&F) are used. They evaluate both the region similarity (Jaccard) and the contour accuracy (F-measure) over a sequence of frames, providing a comprehensive assessment of tracking and segmentation quality over time \cite{cheng2022xmemlongtermvideoobject}.
\end{itemize}

\section{Challenges and Future Research Directions}
\label{sec:challenges}

The integration of Large Language Models (LLMs) with vision-language segmentation (VLSeg) in intelligent transportation systems (ITS) presents several critical challenges that must be addressed to ensure reliable deployment in safety-critical applications. These challenges span computational efficiency, data availability, safety guarantees, and system integration, requiring concerted research efforts to advance the field \cite{cui2023surveymultimodallargelanguage, zhang2024vlmad}. This section also outlines future research directions to tackle these issues.
\begin{itemize}

\item \textbf{Challenge 1: Computational Efficiency for Real-Time Systems}
\label{sub:computational}

Real-time performance is paramount for ITS applications, where decisions must be made within milliseconds to ensure safe navigation and collision avoidance. As discussed in Section \ref{sec:architecture}, the reliance on large transformer-based architectures, particularly with computationally expensive cross-attention mechanisms, means that current VLSeg models like SAM \cite{kirillov2023segment} and SEEM \cite{zou2023seem} face significant latency challenges \cite{elhassan2024real, wang2024autort}.

\textbf{Future Outlook:} Research is focused on advanced model compression, including structured pruning \cite{han2015deepcompression}, knowledge distillation \cite{hinton2015distilling}, and quantization \cite{gholami2022survey} to create lightweight yet powerful models. As noted in our comparative analysis, models like MobileSAM \cite{zhang2023mobilesam} and EdgeViT \cite{pan2022edgevitscompetinglightweightcnns} are direct results of this effort, and work continues on adapting newer, efficient architectures like Mamba for ITS contexts through techniques like unstructured pruning \cite{shihab2025efficient}. Furthermore, hardware-aware Neural Architecture Search (NAS) is being used to design architectures optimized for automotive-grade processors \cite{ benmeziane2021comprehensivesurveyhardwareawareneural}. Future systems will likely employ adaptive fusion strategies that dynamically adjust processing based on scene complexity and available resources \cite{li2024promptad}, potentially using edge-cloud collaboration to distribute computational loads for non-critical tasks \cite{wang2024smartcity}.

\item \textbf{Challenge 2: Data Availability and Open-World Generalization}
\label{sub:data}

The success of VLSeg models depends on high-quality, diverse training data. As noted in Section \ref{sec: datasets}, while foundational datasets like Cityscapes provide excellent benchmarks, they are often scarce for specific ITS scenarios, especially for rare events or "long-tail" objects. This limits the ability of even powerful foundation models to generalize to unseen, open-world conditions.

\textbf{Future Outlook:} To address data scarcity, future work will lean heavily on generative AI and synthetic data pipelines (e.g., LLM-Seg40K \cite{yu2024llmseg}, SynthCity \cite{griffiths2019synthcitylargescalesynthetic}, GAIA \cite{hu2023gaia}) to create vast, diverse, and automatically annotated datasets. While general vision datasets like COCO \cite{lin2014coco} and ADE20K \cite{zhou2017ade20k} have been foundational, the field requires more ITS-specific data. Active learning \cite{settles2009active} and automated annotation frameworks like AutoSeg \cite{zhao2024autoseg} will reduce manual labeling costs. For open-world generalization, the focus is on advancing zero-shot, few-shot, and continual learning methods. Continual learning, in particular, is critical for enabling models to learn from a continuous stream of new driving data without catastrophic forgetting of previously learned knowledge \cite{wang2024continual, delange2021continual}.

\item \textbf{Challenge 3: Safety, Reliability, and Explainability}
\label{sub:safety}

Perhaps the most significant barrier to adoption, and a core perspective of this survey, is the challenge of ensuring safety, reliability, and explainability. Models must be robust against adversarial attacks and perform reliably under adverse conditions. Critically, as foreshadowed in our discussion of end-to-end systems in Section \ref{sec:end_to_end}, their decisions must be interpretable, especially in case of failure. Without a clear ``why" behind an action, true safety is unattainable.

The failure modes for VLSeg in ITS are specific and severe. For example, under extreme weather like heavy snow or rain, cameras can be obscured and LiDAR point clouds can become noisy, leading a model to fail to segment a pedestrian or misclassify a lane marking; this is an active area of research, with methods leveraging generative models to create robust all-weather perception systems \cite{sivaraman2025clearvision}. Sensor degradation, such as a smudged camera lens, can have similar effects. Beyond environmental factors, these models are also vulnerable to adversarial prompts; a malicious actor could potentially craft a text prompt that causes the system to ignore a stop sign or incorrectly segment a clear path. A more subtle failure mode is semantic misinterpretation, where the model correctly segments an object but misunderstands the context—for instance, segmenting a plastic bag floating in the wind as a solid obstacle, causing unnecessary and dangerous braking. Understanding and mitigating these specific failure modes is a critical area of ongoing research \cite{abutami2024using}.

\textbf{Future Outlook:} Research is moving towards formal verification methods and runtime monitoring to provide safety guarantees \cite{katz2017reluplex}. Adversarial training and certified defenses are key areas of research to improve robustness against malicious inputs \cite{madry2017towards}. Frameworks like Multi-Shield \cite{wang2023robust} are exploring multimodal defenses. To address safety in high-risk scenarios and provide end-to-end guarantees, recent models like SafeSeg \cite{li2024safeseg} and VLM-AD \cite{xu2024vlmadendtoendautonomousdriving} integrate predictive reasoning with segmentation to directly inform safer driving decisions. A major future direction, and the one most enabled by LLMs, is explainable AI (XAI). The goal is to create systems that not only perform segmentation but also provide causal reasons for their decisions (e.g., ``The object is segmented as a pedestrian because it has human-like shape and motion"), as explored in \cite{yang2024explainable, guidotti2018survey}. Fail-safe mechanisms, including multi-modal redundancy with LiDAR and radar, will be standard \cite{wang2024multisensor, cui2023surveymultimodallargelanguage}.

\item \textbf{Challenge 4: Integration and Standardization}
\label{sub:integration}

\begin{table*}[!htbp]
\centering
\caption{Practical Deployment Considerations for VLSeg Models in Intelligent Transportation Systems}
\label{tab:deployment_considerations}
\begin{tabular}{|p{2.8cm}|p{3.5cm}|p{4cm}|p{2.5cm}|p{2.7cm}|}
\hline
\textbf{Deployment Aspect} & \textbf{Challenges} & \textbf{Solutions} & \textbf{Example Models/Techniques} & \textbf{ITS Impact} \\
\hline
\textbf{Hardware Constraints} 
& Limited GPU memory and power budget on automotive-grade edge devices. High computational cost of large foundation models. 
& Model compression (quantization, pruning, distillation); hardware-aware Neural Architecture Search (NAS); use of specialized AI accelerators. 
& MobileSAM \cite{zhang2023mobilesam}, EdgeViT \cite{pan2022edgevitscompetinglightweightcnns} 
& Enables on-vehicle deployment for real-time perception without excessive power consumption, crucial for electric vehicles. \\
\hline
\textbf{Real-Time Requirements} 
& High inference latency of complex models (e.g., with cross-attention) exceeds the sub-100ms threshold for safety-critical control loops. 
& Knowledge distillation to smaller, faster student models; exploring efficient architectures (e.g., Mamba, state-space models); dynamic model scaling based on scene complexity. 
& \cite{shihab2025efficient} (from paper), AutoRT 
& Ensures timely detection and segmentation of dynamic obstacles, enabling safe collision avoidance and path planning. \\
\hline
\textbf{Data \& Communication} 
& Continuous stream of high-resolution sensor data; limited bandwidth for Vehicle-to-Everything (V2X) communication. 
& On-device data pre-processing and feature extraction; edge-cloud collaborative systems for offloading non-critical tasks; Federated Learning for privacy-preserving fleet training. 
& Federated Learning \cite{mcmahan2017communication}, Collaborative Perception \cite{chen2019cooper} 
& Allows for fleet-wide model improvement without compromising user privacy and enables vehicles to share perception data to "see" around corners. \\
\hline
\textbf{System Integration \& Safety Standards} 
& Integrating VLSeg modules with existing vehicle control units and software (e.g., AUTOSAR). Lack of certified safety and validation protocols. 
& Adherence to automotive software standards; rigorous validation aligned with safety standards like ISO 26262; development of modular, certifiable components. 
& ISO 26262, SOTIF (ISO 21448) 
& Ensures that VLSeg components are reliable, interoperable, and certifiably safe for mass-market deployment in production vehicles. \\
\hline
\textbf{Robustness \& Long-Term Reliability} 
& Performance degradation over time due to sensor aging, environmental wear (e.g., smudges), or changing road layouts. 
& Multi-modal sensor fusion (camera, LiDAR, Radar) for redundancy; continual learning to adapt to changes; online self-calibration and performance monitoring. 
& Multi-Shield \cite{wang2023robust}, Continual Learning \cite{wang2024continual} 
& Provides consistent and reliable perception throughout the vehicle's lifespan, maintaining safety in all weather and environmental conditions. \\
\hline
\end{tabular}
\end{table*}

Seamless integration of VLSeg models into the broader ITS ecosystem, including vehicle-to-infrastructure (V2I) communication, edge computing devices, and existing vehicle control units, poses a significant engineering challenge. A lack of standardization for evaluation and deployment further complicates this.

\textbf{Future Outlook:} The future lies in developing adaptive and secure V2I communication protocols for collaborative perception \cite{wang2024smartcity}. Edge computing will be essential, requiring efficient resource management and dynamic model selection on vehicular hardware \cite{pan2022edgevitscompetinglightweightcnns}. A key effort will be to establish comprehensive benchmarks and deployment guidelines, aligning with automotive safety standards like ISO 26262 \cite{chen2024traffic}. This will ensure interoperability and certified safety for VLSeg components across different manufacturers and systems.

\item \textbf{Challenge 5: Ethical Considerations and Algorithmic Bias}
\label{sub:ethics}

\begin{table*}[!htbp]
\centering
\caption{Ethical Considerations and Mitigation Strategies for VLSeg in ITS}
\label{tab:ethical_considerations}
\begin{tabular}{|p{2.8cm}|p{3.5cm}|p{3.2cm}|p{4cm}|p{2.2cm}|}
\hline
\textbf{Ethical Issue} & \textbf{Impact on ITS} & \textbf{Example(s) in ITS Context} & \textbf{Mitigation Strategies} & \textbf{References} \\
\hline
\textbf{Algorithmic \& Dataset Bias} 
& Inequitable safety outcomes. Poorer performance for underrepresented groups or non-standard environments leads to higher risk. 
& Models trained primarily on Cityscapes (Germany) fail to recognize pedestrians in different cultural attire or non-Western road signs. 
& Create diverse, globally representative datasets; employ fairness-aware learning algorithms; conduct regular bias audits; use synthetic data for edge cases. 
& \cite{torralba2011unbiased}, \cite{zhang2024fairness} \\
\hline
\textbf{Black-Box Decision-Making} 
& Lack of public trust. Inability to determine the cause of failure in post-accident analysis, which hinders debugging and improvement. 
& An autonomous vehicle makes a sudden, unexplained emergency brake on a highway, and it's impossible to determine if the AI's decision was sound or flawed. 
& Develop and integrate Explainable AI (XAI) methods; generate natural language justifications for actions; use causal reasoning models to show "why" a decision was made. 
& \cite{guidotti2018survey}, \cite{yang2024explainable}, \cite{kim2023lmdrive} \\
\hline
\textbf{Adversarial Vulnerability} 
& Malicious actors can exploit system vulnerabilities to cause accidents or traffic chaos. Safety-critical systems can be intentionally deceived. 
& A physical adversarial patch on a stop sign makes it invisible to the system. A carefully crafted text prompt causes a VLSeg system to ignore cyclists. 
& Adversarial training and data augmentation; certified defense mechanisms; input sanitization for language prompts; multi-modal redundancy (cross-checking camera with LiDAR). 
& \cite{madry2017towards}, \cite{wang2023robust} \\
\hline
\textbf{Automation Bias \& Over-Reliance} 
& Human drivers in semi-autonomous systems (SAE Levels 2-3) become complacent and fail to monitor the system, leading to accidents when the AI fails. 
& A driver using a lane-keeping system stops paying attention and cannot react in time when the system fails to detect a faded lane marking on a curve. 
& Robust driver monitoring systems (DMS); clear Human-Machine Interfaces (HMIs) that communicate system limitations and confidence levels; shared autonomy frameworks. 
& \cite{wang2023voyager} \\
\hline
\textbf{Accountability \& Liability} 
& In case of an accident, it is legally and ethically unclear who is responsible: the owner, manufacturer, software developer, or data provider. 
& A VLSeg model fails to segment a road obstacle due to a rare sensor glitch, causing a crash. Determining legal fault is extremely complex. 
& Development of new legal frameworks (e.g., for product liability); mandating comprehensive "black box" data recorders for all AVs; creating clear industry standards for validation. 
& (Policy-focused legal reviews) \\
\hline
\end{tabular}
\end{table*}

Beyond technical hurdles, the deployment of LLM-augmented segmentation in ITS raises significant ethical questions. The data used to train these models can contain inherent biases, which can lead to inequitable and unsafe outcomes. For example, if a dataset is predominantly collected in one geographic region (e.g., North America), the models may perform worse at recognizing pedestrian behaviors or road signs in other parts of the world. There is a documented ``long-tail" problem where models perform poorly on underrepresented groups \cite{zhang2024fairness}, which in a driving context, could mean a higher risk for certain demographics of pedestrians. Furthermore, the decision-making process of these large models is often opaque (the ``black box" problem), making it difficult to audit or explain failures, which is a major concern for accountability in the event of an accident \cite{yang2024explainable}.

\textbf{Future Outlook:} The ITS research community must prioritize the development of ``fair" and ``transparent" AI. This involves creating more geographically and demographically balanced datasets and developing techniques for bias detection and mitigation \cite{torralba2011unbiased}. Explainable AI (XAI) is a critical research frontier, aiming to create models that can articulate the reasoning behind their predictions (e.g., ``I am slowing down because I have segmented a child running towards the street"). Future regulatory frameworks will likely mandate auditable AI systems in autonomous vehicles, requiring a shift away from purely performance-driven metrics towards a more holistic evaluation that includes fairness, transparency, and ethical alignment.

%\subsection{Long-Term Speculative Directions}
%\label{sub:emerging}

%Looking further ahead, several emerging technologies could reshape VLSeg in ITS. Neuromorphic computing, with its brain-inspired, event-based processing, promises unparalleled energy efficiency for real-time tasks. Hybrid neural-symbolic AI could integrate common-sense reasoning into perception models, allowing them to understand context in a more human-like way. While highly speculative, quantum computing could eventually offer breakthroughs in solving complex optimization problems inherent in large-scale model training and scene understanding. These directions, while not immediately deployable, represent exciting long-term frontiers for the field.

\end{itemize}
\section{Conclusion}

The convergence of Large Language Models and image segmentation has catalyzed a paradigm shift in intelligent transportation systems, moving the field beyond simple perception towards integrated, language-driven cognition. This survey has charted the evolution from classical segmentation methods to promptable, open-vocabulary VLSeg architectures, providing a comprehensive taxonomy, a comparative analysis of state-of-the-art models, and a review of their applications in ITS. We have explored the architectural components, benchmark datasets, and advanced frontiers such as 3D and video segmentation. The analysis underscores a clear trajectory: the future of ITS lies in systems that not only see the world but can also understand, reason about, and communicate their understanding of it.

Significant challenges remain. The computational demands of today's models are often misaligned with the real-time constraints of safety-critical systems. Ensuring reliability in the face of adversarial attacks and adverse conditions is paramount and largely an unsolved problem. Furthermore, the scarcity of large-scale, richly annotated multimodal datasets for the ITS domain continues to be a bottleneck for training and validation. Addressing these issues of efficiency, safety, and data availability is the primary focus of current research.

Looking forward, the integration of VLSeg with end-to-end reasoning and human-in-the-loop systems promises to redefine the relationship between humans and autonomous vehicles. The ultimate goal is to create not just autonomous, but truly collaborative agents that can leverage a deep, language-grounded understanding of their environment to navigate the complexities of real-world driving. These systems will be able to follow complex instructions, explain their decisions in real-time, and request assistance with clarity. Achieving this vision will require continued innovation in efficient model design, robust multi-sensor fusion, and the development of fair and transparent AI. The journey towards this future is well underway, and LLM-augmented segmentation is a foundational pillar upon which it is being built.

\bibliographystyle{IEEEtran}
\bibliography{references}

\end{document}